\documentclass[10pt,journal,compsoc]{IEEEtran}
\usepackage[nocompress]{cite}
\usepackage{url}            % simple URL typesetting
\usepackage{booktabs}       % professional-quality tables
\usepackage{amsfonts}       % blackboard math symbols
\usepackage{nicefrac}       % compact symbols for 1/2, etc.
\usepackage{microtype}      % microtypography
\usepackage{xcolor}         % colors
\usepackage{epsfig}
\usepackage{graphicx}
\usepackage{amsmath}
\usepackage{amssymb}
\usepackage{multirow}
\usepackage{color, colortbl}
\usepackage{subcaption}
\usepackage{algorithm,algpseudocode}
\usepackage{pifont}

\usepackage{xspace}

\makeatletter
\DeclareRobustCommand\onedot{\futurelet\@let@token\@onedot}
\def\@onedot{\ifx\@let@token.\else.\null\fi\xspace}

\makeatother

\usepackage{hyperref}

\usepackage{arydshln}

\begin{document}

% \title{From Sora What Can We See: A survey, taxonomy and future directions for text-to-video generation}
% \title{From Sora What Can We See in Text-to-Video Generation: A survey, taxonomy and future directions}
% \title{Language-guided Video Generation: A survey, taxonomy and future direction}
% \title{LLM-guided Video Generation: A survey, taxonomy and future direction}
% {\footnotesize \textsuperscript{*}Preprint. Work in progress.}
% \title{From Sora What We Can See: A Survey, Challenges, Open Problems and Future Directions of Text-to-Video Generation}
\title{From Sora What We Can See: \\ A Survey of Text-to-Video Generation}

% \author{\textbf{Rui~Sun}\textsuperscript{1}*, \textbf{Yumin~Zhang}\textsuperscript{1}*$^{\dag}$, \textbf{Tejal~Shah}\textsuperscript{1}, \textbf{Jiahao~Sun}\textsuperscript{2}, \textbf{Shuoying~Zhang}\textsuperscript{2}, \textbf{Wenqi~Li}\textsuperscript{1} \\ 
% \textbf{Haoran~Duan}\textsuperscript{1}, 
% \textbf{Bo~Wei}\textsuperscript{1}, \textbf{Rajiv~Ranjan}\textsuperscript{1} \\ %,~\IEEEmembership{Fellow,~IEEE}% <-this % stops a space
% % 
% \textsuperscript{1}Newcastle University \\ 
% \textsuperscript{2}FLock.io

% \IEEEcompsocitemizethanks{
% % \IEEEcompsocthanksitem Rui~Sun, Yumin~Zhang, Tejal~Shah, Bo~Wei and Rajiv~Ranjan are with the School of Computing, Newcastle University, UK. 
% % E-mail: Raj.ranjan@newcastle.ac.uk; ruisun@ieee.org.
% % Jiahao~Sun and Shuoying~Zhang are with FLock.io
% % \IEEEcompsocthanksitem Ling Shao is with Terminus Group, Beijing, China. E-mail: ling.shao@ieee.orgs
% \IEEEcompsocthanksitem * These authors contributed equally to the work.
% \IEEEcompsocthanksitem $^{\dag}$ The corresponding author.
% }
% % \thanks{* Equal contribution.
% %     }
% }

\author{Rui~Sun*, Yumin~Zhang*$^{\dag}$, Tejal~Shah, Jiahao~Sun, Shuoying~Zhang, Wenqi~Li, Haoran~Duan, Bo~Wei, Rajiv~Ranjan \textit{Fellow, IEEE}
\IEEEcompsocitemizethanks{
\IEEEcompsocthanksitem Rui~Sun, Yumin~Zhang, Tejal~Shah, Wenqi~Li, Haoran~Duan, Bo~Wei,Rajiv~Ranjan are with 
Newcastle University, UK.
\protect\\
E-mail: \{ruisun, haoran.duan\}@ieee.org, \{Tejal.Shah, W.Li31, Bo.Wei, Raj.Ranjan\}@newcastle.ac.uk
\IEEEcompsocthanksitem Jiahao~Sun and Shuoying~Zhang are with 
FLock.io, UK.
\protect\\
E-mail: \{sun, shuoying\}@flock.io
% \protect\\
% E-mail: \{jingkang001,kaiyang.zhou,ziwei.liu\}@ntu.edu.sg
% \IEEEcompsocthanksitem 
% Y. Li is with Department of Computer Sciences, University of Wisconsin-Madison, Madison, WI, United States, 53706.
% \protect\\
% E-mail: sharonli@cs.wisc.edu
% }
% \thanks{Manuscript updated April 4, 2023. Discussions, comments, and questions are all welcomed in \href{https://github.com/Jingkang50/OODSurvey/discussions}{https://github.com/Jingkang50/OODSurvey/discussions}.}

% \IEEEcompsocthanksitem Rui~Sun, Yumin~Zhang, Tejal~Shah, Bo~Wei and Rajiv~Ranjan are with the School of Computing, Newcastle University, UK. 
% E-mail: Raj.ranjan@newcastle.ac.uk; ruisun@ieee.org.
% Jiahao~Sun and Shuoying~Zhang are with FLock.io
% \IEEEcompsocthanksitem Ling Shao is with Terminus Group, Beijing, China. E-mail: ling.shao@ieee.org
\IEEEcompsocthanksitem * These authors contributed equally to the work.
\IEEEcompsocthanksitem $^{\dag}$ The corresponding author
(email: Y.Zhang361@newcastle.ac.uk).
}
}

% 

% \and
% \and
% \and
% \and
% \and

\IEEEtitleabstractindextext{%
\begin{abstract}
With impressive achievements made, artificial intelligence is on the path forward to artificial general intelligence. 
Sora, developed by OpenAI, which is capable of minute-level world-simulative abilities can be considered as a milestone on this developmental path.
However, despite its notable successes, Sora still encounters various obstacles that need to be resolved.
In this survey, we embark from the perspective of disassembling Sora in text-to-video generation, and conducting a comprehensive review of literature, trying to answer the question, \textit{From Sora What We Can See}.
Specifically, after basic preliminaries regarding the general algorithms are introduced, the literature is categorized from three mutually perpendicular dimensions: evolutionary generators, excellent pursuit, and realistic panorama. 
Subsequently, the widely used datasets and metrics are organized in detail. 
Last but more importantly, we identify several challenges and open problems in this domain and propose potential future directions for research and development. A comprehensive list of text-to-video generation studies in this survey is available at \href{https://github.com/soraw-ai/Awesome-Text-to-Video-Generation}{https://github.com/soraw-ai/Awesome-Text-to-Video-Generation}
\end{abstract}

}

\maketitle
\IEEEdisplaynontitleabstractindextext
\IEEEpeerreviewmaketitle

% \noindent\includegraphics[width=\textwidth]{diag/performance.pdf}
% \captionof{figure}{xxx}
\begin{figure*}[ht!]
    \centering
    \includegraphics[width=1.0\linewidth]{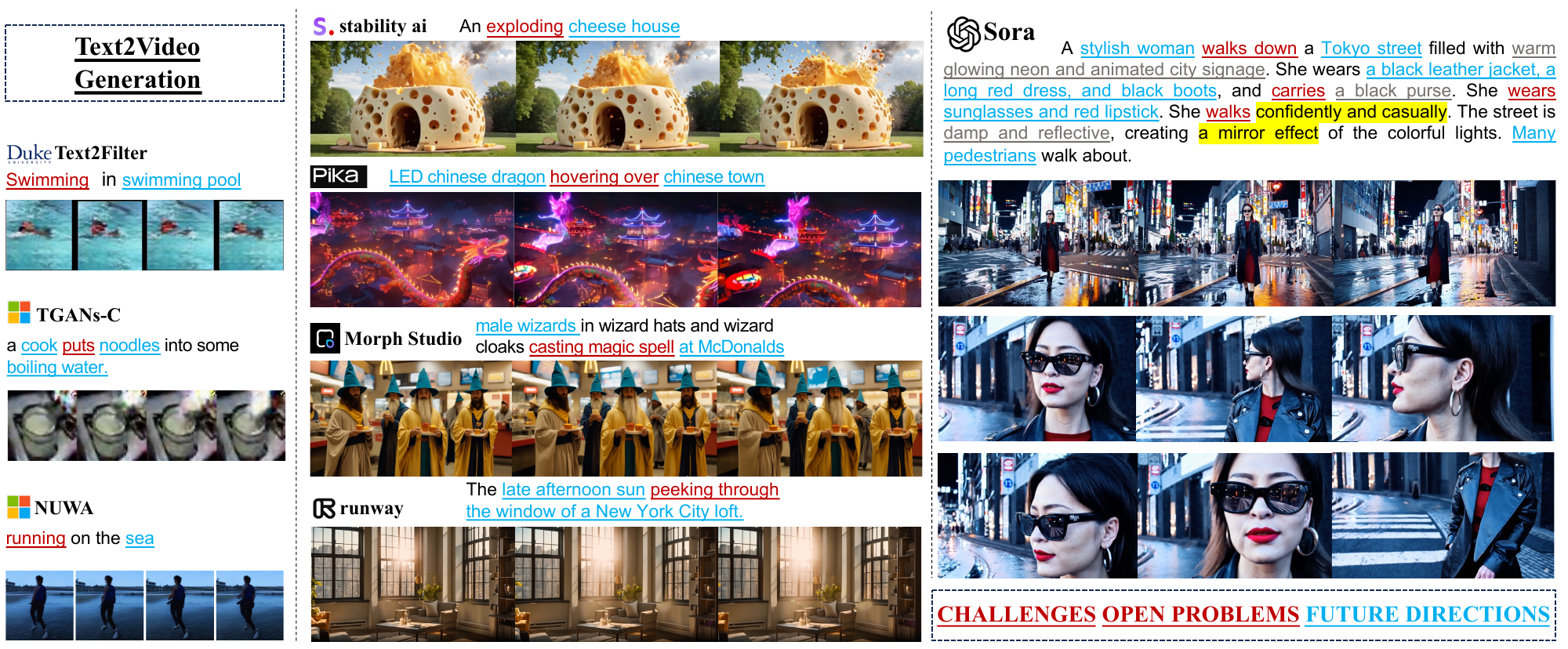}
    \caption{
    Text-to-video (T2V) generation is a flourishing research area, which has gone through several iterations in recent years.
    Early works are limited to simple scenes (low-resolution, single-object, and short-duration).
    Subsequently, benefiting from the success achieved by the diffusion model in the generative area, current works are generating more complex videos and various tools have been commercially successful. 
    Sora, with longer prompts processing capacity and minute-level world-simulative video generation, is an extremely promising T2V tool but it also faces challenges and open problems.
    }
    \label{fig:performance}
\end{figure*}

\section{Introduction}
% 1.AGI -> LLM -> Diffusion model -> Sora -> split key ideas of Sora
% intro 的图可以用一个行业趋势图， 比如说视频生成的demond 走势等等 可以参考这个里面的图https://arxiv.org/pdf/2401.08092.pdf
Recent rapid advancements in the field of AI-generated content (AIGC) mark a pivotal step toward achieving Artificial General Intelligence (AGI), particularly following OpenAI's introduction of their Large Language Model (LLM), GPT-4, in early 2023. AIGC has attracted significant attention from both the academic community and the industry, highlighted by developments such as the LLM-based conversational agent ChatGPT~\cite{chatgpt_web}, and text-to-image~(T2I) models like DALL·E~\cite{betker2023improving}, Midjourney~\cite{midjourney_web} and Stable Diffusion~\cite{rombach2022high}. These achievements have significantly influenced the text-to-video~(T2V) domain, culminating in the remarkable capabilities showcased by OpenAI's Sora~\cite{videoworldsimulators2024}, as shown in Fig.~\ref{fig:performance}.

% Despite existing challenges, the early 2024 debut of Sora by OpenAI has shattered the stalemate in video generation technology, which represents an undoubted milestone. As a text-to-video (T2V) generation model, Sora aims to serve as a world simulator, crafting realistic and imaginative scenes from textual instructions, thus marking a leap forward in the field~\cite{videoworldsimulators2024}.  \tobechange{.........}  At this pivotal moment in the evolving landscape of video generation technology, a critical question arises naturally: \textit{From Sora what we can see in text-to-video generation?}
% 
% Sora, designed to function as a world simulator, generates video with realistic and imaginative scenes from textual instructions~\cite{videoworldsimulators2024}. 

As elucidated in \cite{videoworldsimulators2024}, Sora is engineered to operate as a sophisticated world simulator, crafting videos that render realistic and imaginative scenes derived from textual instructions. Its remarkable scaling capabilities enable efficient learning from internet-scale data, achieved through the integration of the DiT model \cite{peebles2023scalable}, which supersedes the traditional U-Net architecture \cite{ronneberger2015u}. This strategic incorporation aligns Sora with advancements similar to those in GenTron \cite{chen2023gentron}, W.A.L.T \cite{gupta2023photorealistic}, and Latte \cite{ma2024latte}, enhancing its generative prowess. Uniquely, Sora has the capacity to produce minute-long videos of high quality, a feat not yet accomplished by existing T2V studies \cite{chen2020long, ge2022long, villegas2022phenaki, he2022latent, wang2023styleinv, zhuang2024vlogger, zhang2024moonshot, zhang2023adding, sitzmann2020implicit, yu2022generating, skorokhodov2022stylegan, wang2023gen, chen2023seine, fridman2024scenescape, yin2023nuwa, voleti2022mcvd}. It also excels in generating videos with superior resolution and seamless quality, paralleling advancements in existing T2V approaches \cite{blattmann2023align, zhang2023show, bar2024lumiere, tian2021good, jiang2023text2performer, bao2019depth, liu2019deep, niklaus2020softmax, kalluri2023flavr}. While Sora significantly enhances the generation of complex objects, surpassing previous works \cite{liu2023detector, wu2023multi, chen2023videodreamer, ruan2024univg}, it encounters challenges with ensuring coherent motion among these objects. Nonetheless, it is paramount to recognize Sora's superior capability in rendering complex scenes with intricate details, both in subjects and backgrounds, outperforming prior studies focused on complex scene \cite{vondrick2016generating, lin2023videodirectorgpt, lu2023flowzero, long2024videodrafter, fridman2024scenescape} and rational layout generation \cite{gupta2018imagine, lu2023flowzero, lian2023llm}.

Based on our best knowledge, there are two surveys that are relevant to ours: \cite{xing2023survey} and \cite{liu2024sora}. \cite{xing2023survey} encompasses a broad spectrum of topics from video generation to editing, offering a general overview, but focusing on only limited diffusion-based text-to-video (T2V) techniques. Concurrently, \cite{liu2024sora} presents a detailed technical analysis of Sora, providing an elementary survey of related techniques, yet it lacks depth and breadth specifically in the T2V domain. In response, our work seeks to bridge this gap by delivering an exhaustive review of T2V methodologies, benchmark datasets, pertinent challenges and unresolved issues, along with prospective future directions, thereby contributing a more nuanced and comprehensive perspective to the field.

\noindent{\textbf{Contributions:}} 
% In this survey, we provide a comprehensive review focusing on the text-to-video generation domain through an in-depth analysis of Sora, by systematically tracking and summarising recent literature. By extracting the main components of Sora, this survey covers the most representative works in T2V. Relevant fundamental knowledge preliminaries concerning generative models and algorithms are also introduced. And details of all literatured works from algorithm and model they used to the methods generating a high-quality video. Furthermore, we conduct a comprehensive introduction about the T2V datasets and revalent evaluation metrics.
In this survey, we conduct an exhaustive review that focused on the text-to-video generation domain, through an in-depth examination of OpenAI's Sora. We systematically track and summarize the latest literature, distilling the core elements of Sora. This paper also elucidates the foundational concepts, including the generative models and algorithms pivotal to this field. We delve into the specifics of the surveyed literature, from the algorithms and models employed to the techniques employed for producing high-quality videos. Additionally, this survey provides an extensive survey of T2V datasets and relevant evaluation metrics. Significantly, we illuminate the current challenges and open problems in T2V research, proposing future directions based on our insights.

% To the best of our knowledge, this survey represents the first comprehensive review of focusing on the text-to-video generation domain through an in-depth analysis of Sora.

\noindent{\textbf{Sections Structure:}} The paper is organized as follows:
% In Section~\ref{sec:pre}, we will cover background knowledge, including the objective of T2V, foundational models and algorithms. In Section~\ref{sec:from_sora}, we primarily present the overview of all related fields base on Sora what we can see especially focus on T2V generation. In Section~\ref{sec:chall_pro}, we delve into analysis and highlight challenges and open problems from Sora to existing T2V research. In Section~\ref{sec:future}, We give potential future directions by analyzing the existing areas and the points of sora, culminating in our concluding remarks in Section 7~\ref{sec:con}.
In Section~\ref{sec:pre}, we provide a foundational overview, including the objectives of T2V generation, along with the core models and algorithms underpinning this technology. In Section~\ref{sec:from_sora}, we primarily offer an extensive overview of all pertinent fields based on our observations of Sora. In Section~\ref{sec:chall_pro}, we conduct a detailed analysis to underscore the challenges and unresolved questions in T2V research, drawing particular attention to insights gleaned from Sora. Section~\ref{sec:future} is dedicated to outlining prospective future directions, informed by our analysis of existing research and the pivotal aspects of Sora. The paper culminates in Section~\ref{sec:con}, where we present our concluding observations, synthesizing the insights and implications drawn from our comprehensive review.
% On the path forward to achieving Artificial General Intelligence (AGI), the emergence of Sora~\cite{videoworldsimulators2024} is a milestone undoubtedly.
% Recent AGI, Sora~\cite{videoworldsimulators2024}

\section{Preliminaries}
\label{sec:pre}
\subsection{Notations}
Given a collection of $n$ videos and associated text description $\{\mathcal{V}_{i}, \mathcal{T}_{i}\}_{i=1}^{n}$, each video $\mathcal{V}_{i} \in \mathbb{R}^{T\times C\times H\times W}$ contains $k$ frames $\mathcal{V}_{i} = \{f_{i}^{1}, f_{i}^{2}, \cdots, f_{i}^{k}\}$, where $C$ is the number of color bands, $H$ and $W$ are the number of pixels in the height and width of one frame, and $T$ reflects the time dimension.

Guided by the input prompts $\{\mathcal{T}^{*}_{j}\}_{j=1}^{m}$, the goal of text-to-video (T2V) is to generate synthetic videos $\{\mathcal{V}^{*}_{j}\}_{j=1}^{m}$ via the designed generator.

\subsection{Foundational Models and Algorithms}

% \subsubsection{GAN/VAE-based}
\subsubsection{Generative Adversarial Networks~(GAN)}
GAN is an unsupervised machine learning model implemented by a system of two neural networks contesting with each other in a zero-sum game framework~\cite{goodfellow2020generative}. GAN is made up of a generator and a discriminator where the generator's role is to produce data (such as images) that are indistinguishable from real data, while the discriminator's role is to distinguish between the generator's fake and real data~\cite{creswell2018generative, wang2017generative}.

% The overall target optimization function could be formulated as:
% % \begin{equation}
% % \min_{G} \max_{D} V(D, G) = \mathbb{E}_{x\sim p_{data}(x)}[\log D(x)] + \mathbb{E}_{z\sim p_{z}(z)}[\log(1 - D(G(z)))]
% % \end{equation}
% % 
% \begin{equation}
% \max _{D} \min _{G} V(G, D)
% \end{equation}
% where
% \begin{equation}
% V(D, G) = \mathbb{E}_{x\sim p_{\text{data}}(x)}[\log D(x)] + \mathbb{E}_{z\sim p_{z}(z)}[\log(1 - D(G(z)))]
% \end{equation}

% Here, generator~$G$ tries to minimize the probability of the discriminator~$D$ making the correct decision, while the discriminator tries to maximize the probability of identifying the generator's output correctly, $x$ is the real data sampled from data distribution $p_{\text{data}}(x)$, $z$ is sampled from the prior distribution $p_{z}(z)$ such as uniform or Gaussian distribution, and $\mathbb{E}(\cdot)$ represents the expectation. This adversarial process leads to the generator creating data increasingly indistinguishable from real data, as it learns to fool the discriminator.

The optimization objective of a GAN involves a min-max game between two entities: the generator~$G$ and the discriminator~$D$. The overall target optimization function is formulated as:

\begin{equation}
\max _{D} \min _{G} \mathbf{V}(G, D),
\end{equation}
where the value function $\mathbf{V}(D, G)$ is defined as:

\begin{equation}
\mathbb{E}_{x\sim p_{\text{data}}(x)}[\log D(x)] + \mathbb{E}_{z\sim p_{z}(z)}[\log(1 - D(G(z)))].
\end{equation}
In this framework, the generator $G$ aims to minimize the probability that the discriminator $D$ makes a correct decision. Conversely, the discriminator endeavors to maximize its probability of correctly identifying the generator's output. The real data $x$ is sampled from the data distribution $p_{\text{data}}(x)$, and the input to the generator, $z$, is sampled from a prior distribution $p_{z}(z)$, which is typically a uniform or Gaussian distribution. The expectation operator $\mathbb{E}(\cdot)$ represents the expected value over the respective probability distributions. 

This adversarial process involves alternating between optimizing the discriminator to distinguish real data from fake, and optimizing the generator to produce data that the discriminator will classify as real. This training process continues until the generator produces data so close to the real data that the discriminator can no longer distinguish between the two.

% The loss function of discriminator could be formulated as:
% \begin{equation}
% J^{(D)}=-\frac{1}{m} \sum_{i=1}^{m_{\text {real }}} y_{\text {real }}^{(i)} \log \left(D\left(x^{(i)}\right)\right)-\frac{1}{m} \sum_{i=1}^{m_{\text {gen }}}\left(1-y_{\text {gen }}^{(i)}\right) \log \left(1-D\left(G\left(z^{(i)}\right)\right)\right)
% \end{equation}
% where, 

% And the loss function of generator could be formulated as:
% \begin{equation}
% J^{(G)}=\frac{1}{m_{g e n}} \sum_{i=1}^{m_{\text {gen }}} \log \left(1-D\left(G\left(z^{(i)}\right)\right)\right)
% \end{equation}
% Generator
\begin{figure*}[ht!]
    \centering
    \includegraphics[width=1.0\linewidth]{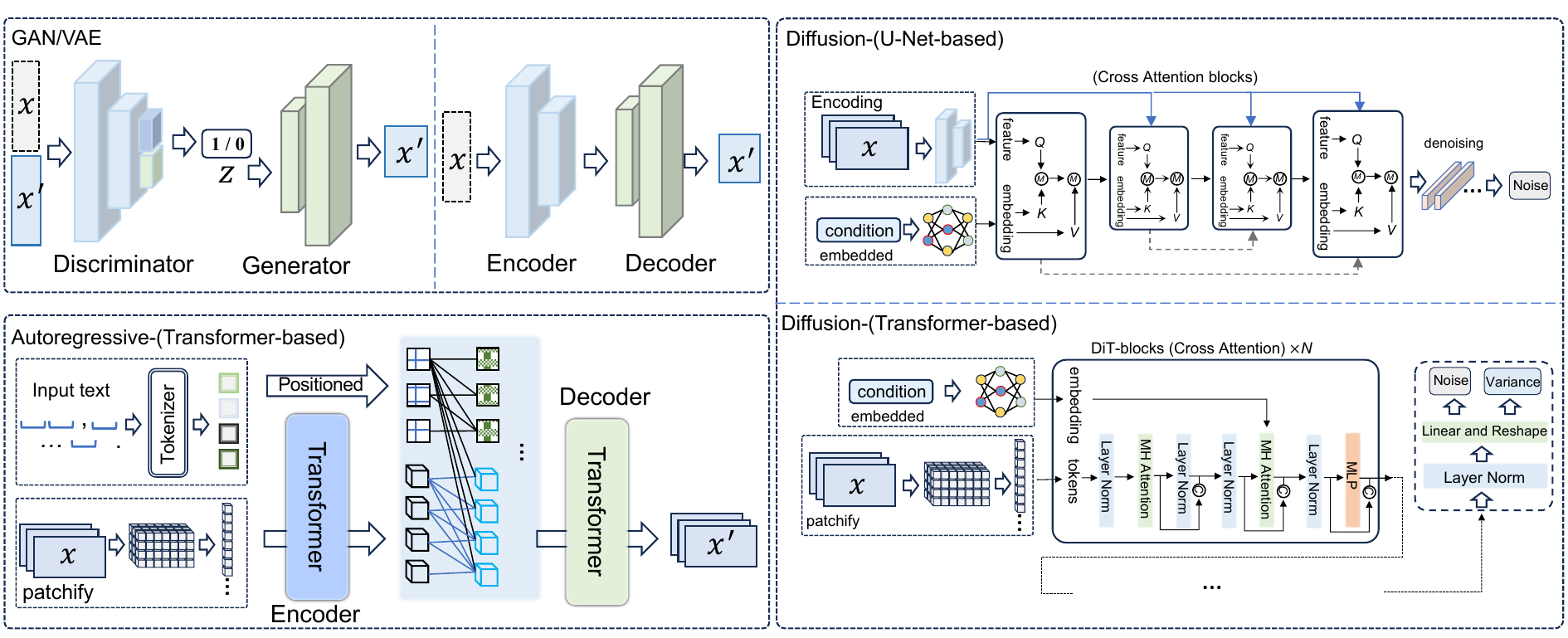}
    \caption{Illustrations of different generators.}
    \label{fig:model_stru}
\end{figure*}

\subsubsection{Variational Autoencoders~(VAE)}
VAE represents a class of deep generative models that are rooted, anchoring their design in the principles of Bayesian inference, which facilitates a structured approach for learning latent representations of data~\cite{kingma2013auto}. A VAE comprises of two primary components: an encoder and a decoder. The encoder, often realized through a neural network, functions by mapping input data to a probabilistic distribution in a latent space. Conversely, the decoder, also implemented as a neural network, aims to reconstruct the input data from this latent representation, thereby enabling the model to capture the underlying data distribution.
% VAE represent a class of deep generative models that are rooted in Bayesian inference and provide a principled framework for learning latent representations of data~\cite{kingma2013auto}. A VAE comprises two main components: an encoder and a decoder. The encoder, typically a neural network, maps the input data to a distribution over a latent space. The decoder, another neural network, reconstructs the input data from the latent space representation.

% $X = \{x^{(i)}\}^N_{i=1}$
Consider a dataset $\mathcal{D} = \{x^{(1)}, x^{(2)}, ..., x^{(N)}\}$ consists of $N$ independent and identically distributed (i.i.d.) samples drawn from a distribution $p(x)$. The objective of a VAE is to approximate this distribution with a parametric distribution model $p_\theta(x)$, where $\theta$ denotes the parameters of the model. This approximation is achieved by introducing a latent variable $\mathbf{z}$, leading to the joint distribution expressed as:
% Consider a dataset $\mathcal{D} = \{x^{(1)}, x^{(2)}, ..., x^{(N)}\}$ , which consists of $N$ independent and identically distributed (i.i.d.) samples drawn from an distribution $p(x)$. The objective of a VAE is to model this distribution with a parametric distribution $p_\theta(x)$, where $\theta$ denotes the parameters of the model. In the context of VAEs, this is achieved by introducing a latent variable $\mathbf{z}$ and defining the model as:

\begin{equation}
p_\theta(x, \mathbf{z}) = p_\theta(x|\mathbf{z})p(\mathbf{z}).
\end{equation}
Here, $p(\mathbf{z})$ is the prior over the latent variables, typically chosen to be a standard Gaussian. The term $p_\theta(x|\mathbf{z})$ represents the likelihood, which is modeled by the decoder network. The encoder facilitates this process by approximating the posterior $p_\theta(\mathbf{z}|x)$ with a variational distribution $q_\omega(\mathbf{z}|x)$, where $\omega$ symbolizes the encoder's parameters.

Training VAE is fundamentally about maximizing the evidence lower bound (ELBO) on the marginal likelihood of the observed data, which is formulated as:
% 
% where $p(\mathbf{z})$ is the prior over the latent variables, typically assumed to be a standard Gaussian, and $p_\theta(x|\mathbf{z})$ is the likelihood, modeled by the decoder network. The encoder network approximates the posterior $p_\theta(\mathbf{z}|\mathbf{x})$ with a variational distribution $q_\omega(\mathbf{z}|x)$, where $\omega$ denotes the encoder parameters.
% 
% The training of VAEs involves optimizing the evidence lower bound (ELBO) on the marginal likelihood of the data:
% 
% where $p(\mathbf{z})$ is the prior over the latent variables, typically assumed to be a standard Gaussian, and $p_\theta(x|\mathbf{z})$ is the likelihood, modeled by the decoder network. The encoder network approximates the posterior $p_\theta(\mathbf{z}|\mathbf{x})$ with a variational distribution $q_\omega(\mathbf{z}|x)$, where $\omega$ denotes the encoder parameters.
% 
% The training of VAEs involves optimizing the evidence lower bound (ELBO) on the marginal likelihood of the data:

\begin{equation}
\mathcal{L}(\theta, \omega; x) = \mathbb{E}_{q_\omega(\mathbf{z}|x)}[\log p_\theta(x|\mathbf{z})] - \text{KL}[q_\omega(\mathbf{z}|x) \| p(\mathbf{z})],
\end{equation}
where the first term is the reconstruction loss, encouraging the decoded samples to match the original inputs, and the second term is the Kullback-Leibler divergence between the approximate posterior and the prior, acting as a regularize. By optimizing the ELBO, VAEs learn to balance the reconstruction fidelity with the complexity of the latent representation, enabling them to generate new samples that are consistent with the observed data.
% where the first term is the reconstruction loss, encouraging the decoded samples to match the original inputs, and the second term is the Kullback-Leibler divergence between the approximate posterior and the prior, acting as a regularizer.

% By optimizing the ELBO, VAEs learn to balance the reconstruction fidelity with the complexity of the latent representation, enabling them to generate new samples that are consistent with the observed data.

% \subsubsection{Diffusion-based}
\subsubsection{Diffusion Model}
Diffusion models~\cite{ho2020denoising} are sophisticated generative models that are trained to create data by inverting a diffusion process, as described by Ho et al. in their 2020 work on denoising. These models incrementally introduce noise to data, eventually converting it into a Gaussian distribution through a specified series of steps. This transformation is represented by a Markov chain comprising latent variables $\{x_t\}_{t=0}^{T}$, where $x_0$ denotes the initial data and $x_T$ represents the fully noised data. The forward diffusion is characterized by a series of transition probabilities $q(x_t|x_{t-1})$ that systematically incorporate noise into the data:
% Diffusion models are a kinds of advanced generative models that learn to generate data by reversing a diffusion process~\cite{ho2020denoising}. The diffusion process gradually adds noise to the data, transforming it into a Gaussian distribution over a fixed number of steps. This process is modeled by a Markov chain of latent variables $\{x_t\}_{t=0}^{T}$, where $x_0$ is the original data, and $x_T$ is completely noisy data. The forward diffusion process is defined by a sequence of transition probabilities $q(x_t|x_{t-1})$ that incrementally add noise to the data:

\begin{equation}
q(x_t|x_{t-1}) = \mathcal{N}(x_t; \sqrt{1-\beta_t}x_{t-1}, \beta_t\mathbf{I}),
\end{equation}
in which $\{\beta_t\}_{t=1}^T$ are predefined small variances that gradually intensify the noise.
% where $\{\beta_t\}_{t=1}^T$ are small variance schedules that progressively increase the noise level.

\textbf{Denoising Diffusion Probabilistic Models (DDPM)~\cite{ho2020denoising}}, highlighted in \cite{ho2020denoising}, are a prominent subset of diffusion models. They interpret the generation process as a backward diffusion that sequentially purifies the data, converting random noise back into a sample from the intended distribution. DDPMs ascertain a denoising function $\mathbf{\epsilon}_\theta(x_t, t)$ to estimate the noise added at each step, with the reverse mechanism defined by:
% DDPMs are a specific type and widely used of diffusion model that frame the generative process as a reverse diffusion process, which aims to denoise the data step-by-step, transforming pure noise into a sample from the target distribution. The model learns a denoising function $\mathbf{\epsilon}_\theta(\mathbf{x}_t, t)$, which predicts the noise added at step $t$, and the reverse process is defined as:

\begin{equation}
p_\theta(x_{t-1}|x_t) = \mathcal{N}(x_{t-1}; \mu_\theta(x_t, t), \Sigma_\theta(x_t, t)),
\end{equation}
where $\mu_\theta(x_t, t)$ and $\Sigma_\theta(x_t, t)$ are functions determined by the neural network that incrementally purify the data. The main objective in training involves enhancing the neural network to diminish the disparity between the noisy data and its denoised version, usually by optimizing the variational lower bound.
% are parameterized by the neural network to gradually denoise the data. The training objective is typically the variational lower bound, which involves optimizing the network to minimize the difference between the noisy data and the denoised estimate.

% The key insight of DDPM is to reverse the diffusion process, which requires estimating the reverse conditional probabilities. By training a neural network to denoise the data at each step, the model can generate high-quality samples from noise by iteratively applying the learned denoising function.
The key insight of DDPM is the inversion of the diffusion process, necessitating the estimation of the reverse conditional probabilities. By training the neural network to eliminate noise at each phase, the model is capable of producing high-quality samples from mere noise through the repetitive application of the learned denoising function.

\begin{figure*}[!ht]
    \centering
    \includegraphics[width=1.0\linewidth]{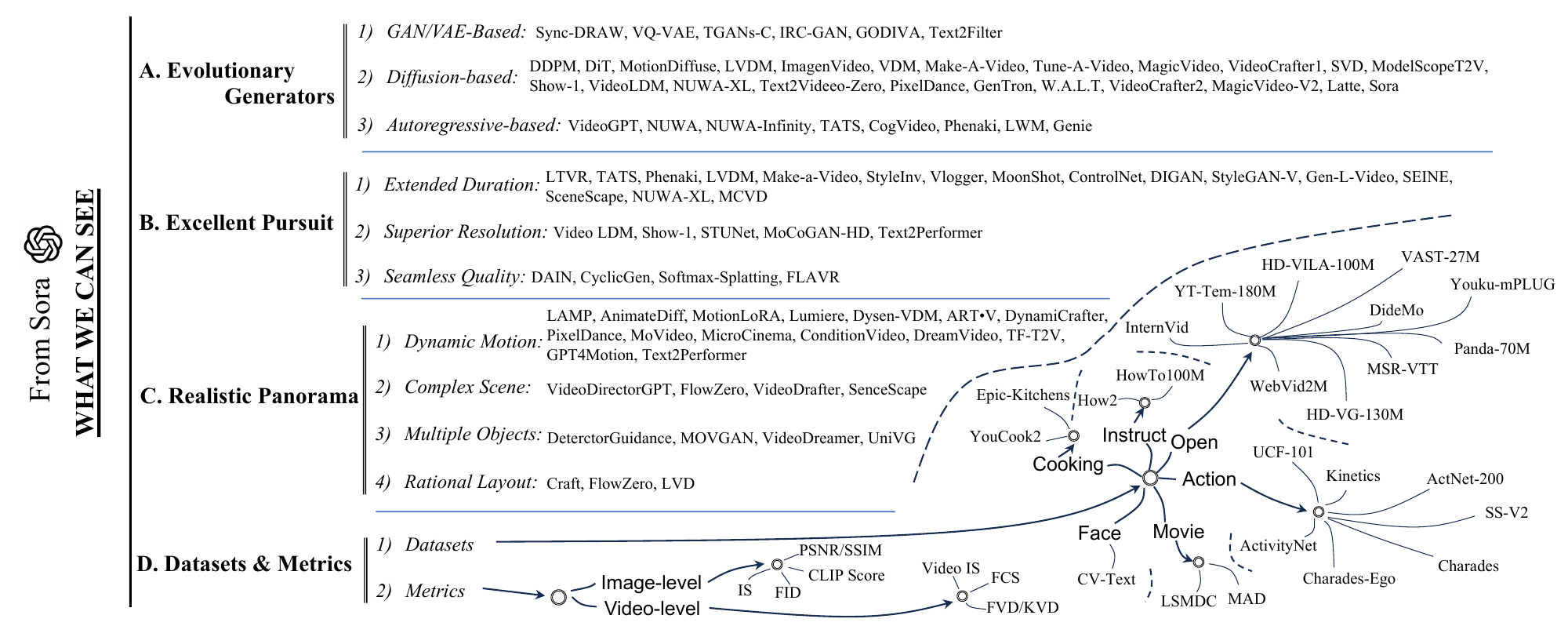}
    \caption{The structure of section \textit{From Sora What We Can See}.}
    \label{fig:structure}
\end{figure*}

% \subsubsection{Autoregressive-based}
\subsubsection{Autoregressive Models}
Autoregressive (AR) models represent a category of statistical models used for understanding and predicting future values in a time series~\cite{ar_defin_wiki}. The fundamental assumption of an AR model is that the current observation is a linear combination of the previous observations plus some noise term. The general form of an AR model of order $o$, denoted as AR($o$), can be expressed as:

\begin{equation}
v_t = \sum_{i=1}^{o} \theta_i v_{t-i} + \epsilon_t,
\end{equation}
where $v_t$ is the current value of the time series. $\theta_1,\theta_2, \ldots, \theta_o$ are the parameters of the model. $v_{t-1}, v_{t-2}, \ldots, v_{t-o}$ are the previous $o$ values of the time series. $\epsilon_t$ is white noise, which is a sequence of uncorrelated random variables with mean zero and a constant finite variance.
% The order $o$ of the model indicates the number of previous time points to be used for predicting the current value. The parameters of the AR model are typically estimated using methods such as the Yule-Walker equations, Maximum Likelihood Estimation, or Least Squares Estimation. Once the model is fitted, it can be used for forecasting future values of the time series.

In the context of neural networks, autoregressive models have been extended to model complex patterns in data. They are fundamental to various architectures, particularly in sequence modeling and generative modeling tasks. By capturing the dependencies between elements in a sequence, autoregressive models can generate or predict subsequent elements in a sequence, making them crucial for applications such as language modeling, time series forecasting, and generative image modeling.

\subsubsection{Transformers}
% focus autoregressive transformer based

% The goal of T2V is to generate synthetic videos $\mathcal{V}^{*}_{j}$ guided by the edited input prompt $\mathcal{T}^{*}_{j}$.

% video $\mathcal{V} = \{v_i \vert i \in [1, m] \}$ contains $m$ frames that can be described by the source prompt $\mathcal{P}$.
% Each video

% The goal of text-to-video generation is to generate a novel video $\mathcal{V}^{*}$ guided by the edited text prompt $\mathcal{P}^{*}$.

The Transformer~\cite{vaswani2017attention} model operates on the principle of self-attention, multi-head attention, and position-wise feed-forward networks.

\textbf{Self-attention mechanism} enables the model to prioritize different parts of the input sequence. It calculates attention scores using the equation:
% \textbf{Self-attention mechanism} allows the model to weigh the importance of different positions in the input sequence. For a single attention head, the attention scores are computed as:

\begin{equation}
    \text{Attention}(Q, K, V) = \text{softmax}\left(\frac{QK^T}{\sqrt{d_k}}\right)V.
\end{equation}
Here, $Q$, $K$, and $V$ correspond to the query, key, and value matrices, which are derived from the input embeddings, with $d_k$ being the key's dimension.

% where $Q$, $K$, and $V$ correspond to the query, key, and value matrices respectively, derived from the input embeddings. $d_k$ denotes the dimension of the key.

% \textbf{Multi-head attention mechanism} extends the model's ability to focus on different positions, by projecting the queries, keys, and values $h$ times with different learned linear projections, and then performing the attention function in parallel. The outputs are then concatenated and linearly transformed. The formulation is given by:
\textbf{Multi-head attention mechanism} enhances the model's capability to attend to various positions by applying different learned projections to the queries, keys, and values $h$ times, executing the attention function concurrently. The results are concatenated and linearly transformed as shown:
\begin{equation}
    \text{MultiHead}(Q, K, V) = \text{Concat}(\text{head}_1, ..., \text{head}_h)\mathcal{W}^O,
\end{equation}
where
\begin{equation}
    \text{head}_i = \text{Attention}(Q\mathcal{W}_i^Q, K\mathcal{W}_i^K, V\mathcal{W}_i^V).
\end{equation}
$\mathcal{W}_i^Q$, $\mathcal{W}_i^K$, $\mathcal{W}_i^V$, and $\mathcal{W}^O$ are the learned parameter matrices.
% Here, $W_i^Q$, $W_i^K$, $W_i^V$, and $W^O$ are parameter matrices that are learned during training.

% \textbf{Position-Wise Feed-Forward Networks}. Each layer in the Transformer contains a fully connected feed-forward network, which is applied to each position separately and identically. This consists of two linear transformations with a ReLU activation in between:

\textbf{Position-Wise Feed-Forward Networks} are included in each layer of the Transformer, applying two linear transformations with a ReLU activation in the middle to each position identically:
\begin{equation}
    \text{FFN}(x) = \max(0, x\mathcal{W}_1 + b_1)\mathcal{W}_2 + b_2.
\end{equation}
The weights and biases of the linear transformations are denoted by $\mathcal{W}_1$, $b_1$, $\mathcal{W}_2$, and $b_2$.

% where $\mathcal{W}_1$, $b_1$, $\mathcal{W}_2$, and $b_2$ are the weights and biases of the linear transformations.

% \textbf{Layer Normalization and Residual Connection}. The Transformer employs residual connections around each of the sub-layers, followed by layer normalization~(LayerNorm). The output of each sub-layer is:
\textbf{Layer Normalization and Residual Connection} are utilized, with the Transformer adding residual connections around each sub-layer, followed by layer normalization. The output is given by:

\begin{equation}
    \text{LayerNorm}(x + \text{Sublayer}(x)),
\end{equation}
where $\text{Sublayer}(x)$ is the function implemented by the sub-layer itself.

% \textbf{Positional Encoding}. Since the model contains no recurrence or convolution, positional encodings are added to the input embeddings to give the model access to token position information. The positional encodings have the same dimension as the embeddings, so that the two can be summed. They are defined as:

\textbf{Positional Encoding} is incorporated into the input embeddings to provide position information, as the model lacks recurrence or convolution. These encodings are summed with the embeddings and are defined as:

\begin{equation}
    PE_{(pos, 2i)} = \sin(pos / 10000^{2i/d_{\text{model}}}),
\end{equation}
\begin{equation}
    PE_{(pos, 2i+1)} = \cos(pos / 10000^{2i/d_{\text{model}}}).
\end{equation}
Here, $pos$ represents the position and $i$ is the dimension.

\section{From Sora What We Can See}
\label{sec:from_sora}
Following significant breakthroughs in text-to-image technology, humans have ventured into the more challenging domain of text-to-video generation, capable of conveying and encapsulating a richer array of visual information. Even though the pace of research in this realm has been gradual in recent years, the launch of Sora has dramatically reignited optimism, marking a momentous shift and breathing new life into the momentum of the field.

Therefore, in this section, we systematically classify the key insights we saw from Sora especially T2V generation area into three main categories and offer detailed reviews for each: \textit{Evolutionary Generators} (see Section \ref{sec:t2v-generators}), \textit{Excellent Pursuit} (see Section \ref{sec:excellent-pursue}), \textit{Realistic Panorama} (see Section \ref{sec:realistic-panorama}) and \textit{Dataset and Metrics} (see Section \ref{sec:dataset-metrics}). The comprehensive structure is illustrated in Fig.~\ref{fig:structure}.

\subsection{Evolutionary Generators}
\label{sec:t2v-generators}
\begin{figure*}[ht!]
    \centering
    \includegraphics[width=\textwidth]{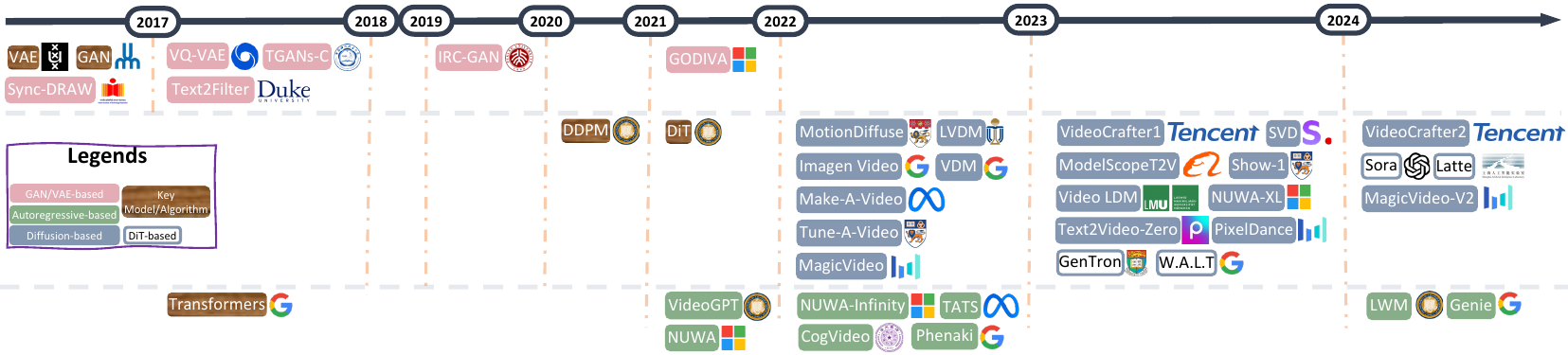}
    \caption{T2V Generators Evolutionary timeline based on foundational algorithms.}
    \label{fig:evo_timeline}
\end{figure*}
The technical report of the cutting-edge Sora~\cite{videoworldsimulators2024} demonstrates that while the video generated from Sora represents a significant leap forward compared to existing works, the advancement in the algorithmic design of text-to-video (T2V) generators has been cautiously incremental. Sora primarily advances the field by refining existing works through intricate splicing and sophisticated optimization techniques. This observation underscores the importance of a comprehensive review of the currently available T2V algorithms. By examining the foundational algorithms underpinning contemporary T2V models, we categorize them into three principal frameworks: GAN/VAE-based, Diffusion-based, and Autoregressive-based.

\subsubsection{GAN/VAE-based}
\label{sec:gan-vae-based}
% 1. VAE
% 2. GAN
In the initial phases of exploration within the text-to-video domain, researchers primarily focused on devising generative models based on neural networks (NN), such as Variational Autoencoders (VAE)~\cite{mittal2017sync,van2017neural,wu2021godiva,li2018video} and Generative Adversarial Networks (GAN)~\cite{li2018video,deng2019irc, pan2017create}. These pioneering efforts laid the groundwork for understanding and developing complex video content directly from textual descriptions.

A pioneering work in automatic video generation from textual descriptions is presented in \cite{mittal2017sync}, where the authors introduced an innovative method that integrates VAE with a recurrent attention mechanism. This approach generates sequences of frames that evolve over time, guided by textual inputs, which uniquely focus on each frame while learning a latent distribution across the entire video. Despite its innovation, the VAE framework faces a notable challenge termed `posterior collapse'. To mitigate this, \cite{van2017neural} introduced the VQ-VAE model, an innovative model that combines the advantages of discrete representation with the efficacy of continuous models. Utilizing vector quantization, VQ-VAE excels in generating high-quality images, videos, and speech. Similarly, GODIWA~\cite{wu2021godiva} is designed based on VQ-VAE architecture pre-trained on HowTo100M~\cite{miech2019howto100m} dataset, showcasing its ability to be fine-tuned on downstream video generation tasks and its remarkable zero-shot capabilities.
% VideoGPT VAE+Transformer

Rather than generating videos directly from textual input, the authors in \cite{pan2017create} proposed TGANs-C, a method that creates videos from textual captions. This approach integrates a novel methodology that includes 3D convolutions and a multi-component loss function, ensuring the videos are both temporally coherent and semantically consistent with the captions. 
% Building on this innovation, 
Expanding upon these innovations, \cite{li2018video} proposed an advanced hybrid model that combines a VAE and GAN, proficiently capturing both the static attributes (e.g., background settings and object layout) and dynamic elements (e.g., movements of objects or characters) from textual descriptions. This model elevates the process of video generation from mere textual inputs to a more complex and nuanced creation of video content based on textual narratives. In a complementary advancement, \cite{deng2019irc} innovatively combines GAN with Long Short-Term Memory (LSTM)~\cite{lstm} networks, significantly improving the visual quality and semantic coherence of text-generated videos, ensuring a closer alignment between the generated content and its textual descriptions.

\subsubsection{Diffusion-based}
\label{sec:diff-based}
% 1. Non-DiT
% 2. DiT
% 讲模型, 怎么设计模型来解决什么问题
In the groundbreaking paper by Ho et al.~\cite{ho2020denoising}, the introduction of diffusion models marked a significant milestone in the text-to-image (T2I) generation domain, leading to the development of pioneering models such as DALL·E~\cite{betker2023improving}, Midjourney~\cite{midjourney_web}, and Stable Diffusion~\cite{blattmann2023stable}. Recognizing that videos fundamentally comprise sequences of images considered with spatial and temporal information, researchers have begun to investigate the potential of diffusion architecture-based models for generating high ﬁdelity videos from textual descriptions.

Video Diffusion Models (VDM)~\cite{ho2022video}, a pioneering advancement in text-to-video generation, significantly enhance the standard image diffusion approach to cater to video data. VDM addresses the critical challenge of temporal coherence in generating high-fidelity videos. By innovating with a 3D U-Net~\cite{cciccek20163d} architecture, the model is tailored for video applications, enhancing traditional 2D convolutions to 3D while integrating temporal attention mechanisms to ensure the generation of temporally coherent video sequences. This design maintains spatial attention and incorporates temporal dynamics, enabling the model to generate coherent video sequences. In a similar vein, MagicVideo~\cite{zhou2022magicvideo} and its successor, MagicVideo-V2~\cite{zhou2022magicvideo}, are significant innovations in text-to-video generation, leveraging latent diffusion models to tackle challenges like data scarcity, complex temporal dynamics, and high computational costs. MagicVideo introduces a 3D U-Net-based architecture with enhancements like a video distribution adaptor and directed temporal attention, enabling efficient, high-quality video generation that maintains temporal coherence and realism. It operates in a latent space, focusing on keyframe generation and efficient video synthesis. MagicVideo-V2 builds upon this, incorporating a multi-stage pipeline with modules for Text-to-Image, Image-to-Video, Video-to-Video, and Video Frame Interpolation. 

% Latent Diffusion
Latent space exploitation is a common theme among several models. LVDM~\cite{he2022latent} proposed a hierarchical latent video diffusion model that compresses videos into a lower-dimensional latent space, enabling efficient long video generation and reducing computational demands. The conventional framework is restricted to producing short videos, the length of which is predetermined by the number of input frames provided during the training phase. To overcome this limitation, the LVDM introduces a conditional latent diffusion approach, enabling the generation of future latent codes based on previous ones in an autoregressive fashion. Show-1~\cite{zhang2023show}, PixelDance~\cite{zeng2023make}, and SVD~\cite{blattmann2023stable} leverage both pixel-based and latent-based techniques for generating high-resolution videos. Show-1 starts with pixel-based VDMs for generating accurate, low-resolution keyframes and then uses latent-based VDMs for efficient high-resolution video enhancement, capitalizing on the strengths of both approaches to ensure high-quality, computationally efficient video outputs. PixelDance is built on a latent diffusion framework trained to denoise perturbed inputs within the latent space of a pre-trained VAE, aiming to minimize computational demands. The underlying structure is a 2D UNet diffusion model, enhanced to a 3D variant by incorporating temporal layers, including 1D convolutions and attentions along the temporal dimension, adapting it for video content while maintaining its efficacy with image inputs. This model, capable of joint training with images and videos, ensures high-fidelity outputs by effectively integrating spatial and temporal resolutions. SVD  incorporates temporal convolution and attention layers into a pre-trained diffusion architecture, allowing it to capture dynamic changes over time efficiently. Tune-A-Video~\cite{wu2023tune} extends these concepts by incorporating temporal self-attention to capture consistencies across frames, optimizing computational resources. Specifically, Tune-A-Video aims to solve the computational expense problem and extends a 2D Latent Diffusion Model (LDM) to the spatio-temporal domain to facilitate T2V generation. Researchers in~\cite{wu2023tune} innovated by adding a temporal self-attention layer to each transformer block within the network, enabling the model to capture temporal consistencies across video frames. This design is complemented by a sparse spatio-temporal attention mechanism and a selective tuning strategy that updates only the projection matrices in attention blocks, optimizing for computational efficiency and preserving the pre-learned features of the T2I model while ensuring temporal coherence in the generated video.
% 
% specifically using a variant of the 2D UNet architecture extended to 3D to accommodate temporal dimensions, integrating image instructions for both the initial and final frames alongside text instructions. This innovative approach enables the generation of high-dynamic videos with complex scenes and detailed motions. 

In the realm of video enhancement, VideoLCM~\cite{wang2023videolcm} model is designed as a latent consistency model optimized with a consistency distillation strategy, focusing on reducing the computational burden and accelerating training. It leverages large-scale pre-trained video diffusion models to enhance training efficiency. The model applies the DDIM~\cite{song2020denoising} solver for estimating the video output and incorporates classifier-free guidance to synthesize high-quality content, allowing it to achieve fast and efficient video synthesis with minimal sampling steps. VideoCrafter2~\cite{chen2024videocrafter2} model is distinctively engineered to enhance the spatial-temporal coherence in video diffusion models, employing an innovative data-level disentanglement strategy that meticulously separates motion aspects from appearance characteristics. This strategic design facilitates a targeted fine-tuning process with high-quality images, aiming to substantially elevate the visual fidelity of the generated content without compromising the precision of motion dynamics. This approach builds upon and significantly refines the groundwork laid by VideoCrafter1~\cite{chen2023videocrafter1}, which is structured as a Latent Video Diffusion Model (LVDM), incorporating a video Variational Autoencoder (VAE) and a video latent diffusion process. The video VAE compresses the video data into a lower-dimensional latent representation, which is then processed by the diffusion model to generate videos.

% Gen-1~\cite{esser2023structure} a novel latent video diffusion models for text-to-video synthesis. Gen-1 is a structure and content-guided video diffusion model that edits videos based on textual or visual descriptions. It uniquely integrates depth estimates to represent video structure and CLIP image embeddings for content, enabling fine-grained control over the editing process while maintaining the original video's structural integrity. The model's architecture features temporal layers and a joint training strategy on both images and videos, enhancing its temporal consistency capabilities.

Models like Make-A-Video~\cite{singer2022make} and Imagen Video~\cite{ho2022imagen} extend text-to-image technologies to the video domain. Make-A-Video model leverages advancements in T2I technology and extends them into the video domain without needing paired text-video data. It is designed around three core components: a T2I model, spatiotemporal convolution and attention layers, and a frame interpolation network. Initially, it leverages a T2I model trained on text-image pairs, then extends this with novel spatiotemporal layers that incorporate temporal dynamics, and finally, employs a frame interpolation network to enhance the frame rate and smoothness of the generated videos. Imagen Video employs a sophisticated cascading architecture of video diffusion models specifically tailored for T2V synthesis. This design intricately combines base video diffusion models with subsequent stages of spatial and temporal super-resolution models, all conditioned on textual prompts, to progressively enhance the quality and resolution of generated videos. The model's innovative structure enables it to produce videos that are not only high in fidelity and resolution but also exhibit strong temporal coherence and alignment with the descriptive text. MotionDiffuse~\cite{zhang2024motiondiffuse}: text-driven human motion generation, particularly the need for diversity and fine-grained control in generated motions. utilizing a diffusion model, specifically tailored with a Cross-Modality Linear Transformer for integrating textual descriptions with motion generation. It enables fine-grained control over the generated motions, allowing for independent control of body parts and time-varied sequences, ensuring diverse, realistic outputs. Text2Video-Zero~\cite{khachatryan2023text2video} is built upon the Stable Diffusion T2I model, tailored for zero-shot T2V synthesis. The core enhancements include introducing motion dynamics to the latent codes for temporal consistency and employing a cross-frame attention mechanism that ensures the preservation of object appearance and identity across frames. These modifications enable the generation of high-quality, temporally consistent video sequences from textual descriptions without additional training or fine-tuning, leveraging the existing capabilities of pre-trained T2I models.

NUWA-XL~\cite{yin2023nuwa} introduces a novel ``Diffusion over Diffusion" architecture designed for generating extremely long videos. The primary challenge it addresses is the inefficiency and quality gap in long video generation from existing method, NUWA-XL's architecture innovatively employs a ``coarse-to-fine" strategy. It starts with a global diffusion model that generates keyframes outlining the video's coarse structure. Subsequently, local diffusion models refine these keyframes, filling in detailed content between them, enabling the system to generate videos with both global coherence and fine-grained details efficiently. 

Differing from the approach of fine-tuning pre-trained models, Sora aims at the more ambitious task of training a diffusion model from scratch, presenting a significantly greater challenge. Drawing inspiration from the scalability of transformer architectures, OpenAI has integrated the DiT~\cite{peebles2023scalable} framework into their foundational model architecture, OpenAI shifts the diffusion model from the conventional U-Net~\cite{ronneberger2015u} to a transformer-based structure, harnessing the Transformer's scalable capabilities to efficiently train massive amounts of data and tackle complex video generative tasks. 
% 
% Similarly, aiming to enhance training efficiency through the integration of transformers and diffusion models, GenTron~\cite{chen2023gentron} a DiT model employs strategies like Fully Sharded Data Parallel (FSDP) and activation checkpointing (AC) to optimize GPU memory usage, facilitating the training of large models effectively. Concurrently, W.A.L.T~\cite{gupta2023photorealistic} a DiT model improved training efficiency by employing a window attention architecture for spatial and spatiotemporal modeling, significantly reducing computational demands, and by innovating with AdaLN-LoRA to decrease the number of trainable parameters without compromising performance. They also optimized the noise schedule to better match the training and inference phases, especially beneficial for video generation due to high temporal redundancy. Nonetheless, it is noteworthy that both GenTron and W.A.L.T primarily concentrate on fine-tuning pre-trained models.
% 
Similarly, aiming to enhance training efficiency through the integration of transformers and diffusion models, GenTron~\cite{chen2023gentron} builds upon the DiT-XL/2 structure, transforming latent dimensions into non-overlapping tokens processed through transformer blocks. The model's innovation lies in its text conditioning, employing both adaptive layernorm and cross-attention mechanisms for integrating text embeddings and enhancing interaction with image features. A significant aspect of GenTron is its scalability; the GenTron-G/2 variant expands the model to over 3 billion parameters, focusing on the depth, width, and MLP width of transformer blocks. Concurrently, W.A.L.T~\cite{gupta2023photorealistic} is based on a two-stage process incorporating an autoencoder and a novel transformer architecture. Initially, the autoencoder compresses both images and videos into a lower-dimensional latent space, enabling efficient training on combined datasets. The transformer employs window-restricted self-attention layers, alternating between spatial and spatiotemporal attention, which significantly reduces computational demands while supporting joint image-video processing. This structure facilitates the generation of high-resolution, temporally consistent videos from textual descriptions, showcasing an innovative approach in T2V synthesis. Latte~\cite{ma2024latte} further extends these innovations by employing a series of Transformer blocks to process latent space representations of video data, which are obtained using a pre-trained variational autoencoder. This approach allows for the effective modeling of the complex distributions inherent in video data, handling both spatial and temporal dimensions innovatively.

\subsubsection{Autoregressive-based}
% [arxiv 2021] ---VideoGPT-- VideoGPT: Video Generation using VQ-VAE and Transformers [PDF, code ]
% [ECCV 2022; Microsoft] ---NÜWA-- NÜWA: Visual Synthesis Pre-training for Neural visUal World creAtion [PDF, code ]
% [NIPS 2022; Microsoft] ---NÜWA-Infinity-- NUWA-Infinity: Autoregressive over Autoregressive Generation for Infinite Visual Synthesis [PDF, code ]
% [Arxiv 2020; Tsinghua] ---CogVideo-- CogVideo: Large-scale Pretraining for Text-to-Video Generation via Transformers [PDF, code ]
% *[ECCV 2022] ---TATS-- Long Video Generation with Time-Agnostic VQGAN and Time-Sensitive Transformer [PDF, code]
% *[arxiv 2022; Google] ---PHENAKI-- PHENAKI: VARIABLE LENGTH VIDEO GENERATION FROM OPEN DOMAIN TEXTUAL DESCRIPTIONS [PDF, code ]
% [arxiv 2023.11]Optimal Noise pursuit for Augmenting Text-to-Video Generation [PDF]
% [arxiv 2024.01]WorldDreamer: Towards General World Models for Video Generation via Predicting Masked Tokens [PDF,Page]
% , and Autoregressive Models (ARM)~\cite{.}. 
% Apart from mainstream research on diffusion-based, several studies proposed transformer~\cite{yan2021videogpt} based models in text-to-video generation due to transformers' superior efficiency in handling sequential data and their proven scalability, which is crucial for modeling complex temporal dynamics and high-dimensional data in video generation tasks. Additionally, transformers facilitate direct integration with existing language models, enabling more coherent and contextually rich multimodal outputs.
Recent advancements in T2V generation have prominently featured autoregressive-based transformers as well, recognized for their superior efficiency in handling sequential data and scalability. These attributes are pivotal for modeling the intricate temporal dynamics and high-dimensional data characteristics of video generation tasks. Transformers~\cite{yan2021videogpt} are particularly advantageous due to their seamless integration with existing language models, which fosters the creation of coherent and contextually enriched multimodal outputs.
% 
% \tobechange{basic AR add ART`V paper,  scaling AR for T2I gen Parti~\cite{yu2022scaling}, autoregressive video
% generation models~(ARVM)~\cite{weissenborn2019scaling}}
A notable development in this area is NUWA~\cite{wu2022nuwa}, which integrates a 3D transformer encoder-decoder framework with a specialized 3D Nearby Attention mechanism, enabling efficient and high-quality synthesis of images and videos by processing data across 1D, 2D, and 3D dimensions, showcasing remarkable zero-shot capabilities. Building on this, NUWA-Infinity~\cite{wu2022nuwa-inf} introduces an innovative autoregressive over autoregressive framework, adept at generating variable-sized, high-resolution visuals. It combines a global patch-level with a local token-level model, enhanced by a Nearby Context Pool and an Arbitrary Direction Controller, to ensure seamless, flexible, and efficient visual content generation. 

Further extending these capabilities, Phenaki~\cite{villegas2022phenaki} extends the paradigm with its unique C-ViViT encoder-decoder structure, focusing on the generation of variable-length videos from textual inputs. This model efficiently compresses video data into a compact tokenized representation, facilitating the production of coherent, detailed, and temporally consistent videos. Similarly, VideoGPT~\cite{yan2021videogpt} is a model that innovatively combines VQ-VAE and Transformer architectures to tackle video generation challenges. It employs VQ-VAE to learn downsampled discrete latent representations of videos through 3D convolutions and axial attention, creating a compact and efficient representation of video content. These learned latents are subsequently modeled autoregressively with a Transformer, enabling the model to capture the complex temporal and spatial dynamics of video sequences. 

The Large World Model (LWM)~\cite{liu2024world} represents another stride forward, designed as an autoregressive transformer to process long-context sequences, blending video and language data for multimodal understanding and generation. Key to its design is the RingAttention mechanism, which addresses the computational challenges of handling up to 1 million tokens efficiently, minimizing memory costs while maximizing context awareness. The model incorporates VQGAN for video frame tokenization, integrating these tokens with text for comprehensive sequence processing. On the other hand, Genie~\cite{bruce2024genie} model is crafted as a generative interactive environment tool, which incorporates spatiotemporal (ST) transformers across all its components, utilizing a novel video tokenizer and a causal action model to extract latent actions, which are then passed to a dynamics model. This dynamics model autoregressively predicts the next frame, employing an ST-transformer architecture that balances model capacity with computational efficiency. The model's design leverages the strengths of transformers, particularly in handling the sequential and spatial-temporal aspects of video data, to generate controllable and interactive video environments.

TATS~\cite{ge2022long} is specifically designed for generating long-duration videos, addressing the challenge of maintaining high quality and coherence over extended sequences. The model architecture innovatively combines a time-agnostic VQGAN, which ensures the quality of video frames without temporal dependence, and a time-sensitive transformer, which captures long-range temporal dependencies. This dual approach enables the generation of high-quality, coherent long videos, setting a new standard in the field of video synthesis. 

% WorldDreamer~\cite{wang2024worlddreamer} model is designed to capture general world dynamics, addressing the limitations of existing models that are confined to specific scenarios and struggle with the complexities of diverse environments. WorldDreamer's architecture is based on a Spatial Temporal Patchwise Transformer (STPT), which is designed to understand the dynamics of visual world scenes comprehensively. It leverages a Transformer-based framework, integrating multi-modal prompts for interaction within the world model. This approach allows for efficient video generation, significantly enhancing the model's ability to capture complex motion and physics in various environments. 

CogVideo~\cite{hong2022cogvideo} integrates a multi-frame-rate hierarchical training approach, adapting from a pretrained T2I model, CogView2~\cite{ding2022cogview2}, to enhance T2V synthesis. This design inherits the text-image alignment knowledge from CogView2, using it to generate key frames from text and then interpolating intermediate frames to create coherent videos. The model's dual-channel attention mechanism and recursive interpolation process allow for detailed and semantically consistent video generation.

% Introduce Sora DiT last
% 
% \subsection{Physical Realism}
% % Origin: Physical World Simulation
% \noindent{\textbf{Motion}}

% \noindent{\textbf{Layout}}

% \noindent{\textbf{Causal understanding}}

% \noindent{\textbf{Reflection}}

% \subsection{Interactive Dynamics}
% % Origin: Human Interactive Generation
% \noindent{\textbf{Text-guided spatial details understanding}}

% \subsection{Scene Complexity}
% % Origin: Complex Scenes Generation
% \noindent{\textbf{Multiple Characters}}
% \noindent{\textbf{Multiple Shots}}

\subsection{Excellent Pursuit}
\label{sec:excellent-pursue}
Sora has shown its excellent video generation abilities from the presented demos with long-term duration, high resolution, and smoothness~\cite{sora_web}. 
Based on such merits, we categorized current works into three streams: \textit{extended duration}, \textit{superior resolution}, and \textit{seamless quality}.

\subsubsection{Extended Duration}
Compared with short video generation or image generation tasks, long-term video generation is more challenging as the latter requires the ability to model long-range temporal dependence and maintain temporal consistency with many more frames~\cite{videoworldsimulators2024}. 

Specifically, with the duration extended of generative video, one obstacle is that the prediction error will accumulate.
To address such a challenge, the retrospection mechanism (LTVR)~\cite{chen2020long} is introduced to push the retrospection frames to be consistent with the observed frames, and thus alleviate the accumulated prediction error.
TATS~\cite{ge2022long} achieves the goal of long video generation by combining a time-agnostic VQGAN for high-quality frame generation and a time-sensitive transformer for capturing long-range temporal dependencies. 
Phenaki~\cite{villegas2022phenaki} is presented for generating videos from open-domain textual descriptions. 
By incorporating the designed casual attention, it is capable of working with variable-length videos and generating longer sequences by extending the video with new prompts.
LVDM~\cite{he2022latent} proposed a hierarchical framework for generating longer videos, enabling the production of videos with over a thousand frames.
Through leveraging T2I models for visual learning and unsupervised video data for motion understanding, Make-A-Video~\cite{singer2022make} can generate long videos with high fidelity and diversity without the need for paired text-video data. 
StyleInV~\cite{wang2023styleinv} is capable of generating long videos by leveraging the sparse training approach and the efficient design of the motion generator. 
By constraining the generation space with the initial frame and employing temporal style codes, the method achieves high single-frame resolution and quality, as well as temporal consistency over long durations.
More recently, through enhancing spatial-temporal coherence in the snippet, Vlogger~\cite{zhuang2024vlogger} successfully generates over 5-minute vlogs from open-world descriptions without losing video coherence regarding the script and actors.
MoonShot~\cite{zhang2024moonshot} leverages the Multimodal Video Block that integrates spatial-temporal layers with decoupled cross-attention for multimodal conditioning, and the direct use of pre-trained Image ControlNet~\cite{zhang2023adding} for precise geometry control, and thus to generate long videos with high visual quality and temporal consistency, efficiently handling diverse generative tasks.

Additionally, inspired by the success of Implicit Neural Representations (INR)~\cite{sitzmann2020implicit} in model complex signals, various works have introduced INR in video generation.
For example, DIGAN~\cite{yu2022generating} synthesizes long videos of high resolution without demanding extensive resources for training by leveraging the compactness and continuity of Implicit Neural Representations (INRs), allowing for efficient training on long videos.
Similarly, StyleGAN-V~\cite{skorokhodov2022stylegan} can efficiently produce videos of any length and at any frame rate by treating videos as continuous signals. 
Besides, various works from the perspective of decomposing to address the long-range consistency during the process of video generation.
In Gen-L-Video~\cite{wang2023gen}, the long videos are first perceived as short clips. 
The bidirectional cross-frame attention is developed to mutual the influence among different video clips and thus facilitate finding a compatible denoising path for the long video generation.
SEINE~\cite{chen2023seine} adopts a similar approach that views long videos as compositions of various scenes and shot-level videos of different lengths.
By synthesizing long-term videos of various scenes, SceneScape~\cite{fridman2024scenescape} is proposed to generate long-term videos from text descriptions, addressing the challenge of 3D consistency in video generation.
NUWA-XL~\cite{yin2023nuwa} offers a novel solution based on a coarse-to-fine strategy, developing a ``Diffusion over Diffusion" architecture that enables efficient and coherent generation of extremely long videos.
MCVD~\cite{voleti2022mcvd} generates videos of arbitrary lengths by autoregressively generating blocks of frames, allowing for the production of high-quality frames for diverse types of videos, including long-duration content.

\subsubsection{Superior Resolution}

Compared to low-quality videos, high-resolution videos undoubtedly have a broader potential application such as a simulation engine in the context of autonomous techniques.

On the other hand, high-resolution video generation also poses a challenge to the computational resources.
Considering such a challenge, Video Latent Diffusion Models (LDM)~\cite{blattmann2023align} introduce the off-the-shelf pre-trained image LDM into the video generation while avoiding excessive computing demands.
By training a temporal alignment model, Video LDMs can generate videos up to 1280$\times$2048 resolution.

However, LDMs struggle to generate a precise text-video alignment.
In Show-1~\cite{zhang2023show}, the advantages of pixel-based and latent-based Video Diffusion Models (VDM) are combined as a hybrid model.
Show-1 utilizes the pixel-based VDMs to init the low-resolution video generation and then uses latent-based VDMs for upscaling to get high-resolution videos (up to 572$\times$320).
Recently, STUNet~\cite{bar2024lumiere} adopts a Spatial Supper-Resolution (SSR) model to upsample the output from the base model.
Specifically, to avoid temporal boundary artifacts and ensure smooth transitions between temporal segments, Multi-Diffusion is employed along the temporal axis, which allows the SSR network to operate on short segments of the video, averaging predictions of overlapping pixels to produce a high-resolution video. 
From another perspective, video generation is regarded as a trajectory of discovering the problem in MoCoGAN-HD~\cite{tian2021good} which presents a framework that utilizes contemporary image generators to render high-resolution videos (up to 1024$\times$1024).
In the text-driven human video generation tasks, a subfield of video generation, Text2Performer~\cite{jiang2023text2performer} achieves the goal of generating high-resolution human videos (up to 512$\times$256) by decomposing the latent space for separate handling of human appearance and motion and utilizing continuous pose embeddings for motion modeling.
Such a method ensures the appearance is consistently maintained across frames while producing temporally coherent and flexible human motions from text descriptions.

\subsubsection{Seamless Quality}
In terms of viewability, the high-frame-rate video is much more appealing to viewers compared to the unsmooth, as it avoids common artifacts, such as temporal jittering and motion blurriness~\cite{bao2019depth}.

Depth-aware video frame interpolation method (DAIN)~\cite{bao2019depth} is introduced to exploit depth information to detect occlusion and prioritize closer objects during interpolation.
By involving a depth-aware flow projection layer that synthesizes intermediate flows by sampling closer objects preferentially over farther ones, the method can address the challenge of occlusion and motion in the video frames.
Through leveraging cycle consistency loss along with motion linearity loss and edge-guided training, CyclicGen~\cite{liu2019deep} achieves superior performance in generating high-quality interpolated frames, which is essential for high frame-rate video generation.
Softmax splatting is introduced in Softmax-Splatting~\cite{niklaus2020softmax} to interpolate frames at any desired temporal position, effectively contributing to the generation of high frame-rate videos.
Currently, FLAVR~\cite{kalluri2023flavr} addresses the challenge of generating high frame rate videos through a designed architecture that leverages 3D spatio-temporal convolutions for motion modeling.
Besides, to alleviate the computational limitation, FLAVR directly learns motion properties from video data, which simplifies the training and deployment process.

\subsection{Realistic Panorama}
\label{sec:realistic-panorama}
A key challenge in T2V generation is achieving realistic video output. 
Addressing this involves focusing on the integration of elements essential for realism.
Breaking down a realistic panorama T2V generation, we identify the following key components that should be considered: 1. Dynamic Motion 2. Complex Scene 3. Multiple Objects 4. Rational Layout.
% Following are the key components for Realistic Panorama T2V generation, 1. Dynamic Motion 2. Complex Scene 3. Multiple Objects 4. Rational Layout.

\subsubsection{Dynamic Motion}
% model transfer T2I to T2V
In recent years, although there has been significant advancement in T2I generation, numerous researchers have begun exploring the extension of T2I models to T2V generation, such as LAMP~\cite{wu2023lamp} and AnimateDiff~\cite{guo2023animatediff}. Motion represents one of the critical distinctions between videos and images, and it is a key focus in this evolving field of study~\cite{liang2023movideo}. LAMP focuses on learning motion patterns from a limited dataset. It employs a first-frame-conditioned pipeline that leverages a pre-trained T2I model to create the initial frame, enabling the video diffusion model to concentrate on learning the motion for subsequent frames. This process is enhanced with temporal-spatial motion learning layers that capture both temporal and spatial features, streamlining the motion generation in videos. AnimateDiff integrates a pre-trained motion module into personalized T2I models, enabling smooth, content-aligned animation production. This module is optimized using a novel strategy that learns motion priors from video data, focusing on dynamics rather than pixel details. A key innovation is MotionLoRA, a fine-tuning technique that adapts this module to new motion patterns, enhancing its versatility for varied camera movements. 

% motion coherent and consistency
Motion consistency and coherence are also key challenges in motion generation. Unlike traditional models that generate keyframes and then fill in gaps, often leading to inconsistencies, Lumiere~\cite{bar2024lumiere} uses a Space-Time U-Net architecture to generate the entire video in one pass. This method ensures global temporal consistency by incorporating spatial and temporal down- and up-sampling, significantly improving motion generation performance. Dysen-VDM~\cite{fei2023empowering} integrates a Dynamic Scene Manager (Dysen) that operates in three orchestrated steps: extracting key actions from text, converting these into a dynamic scene graph (DSG), and enriching the DSG with contextual scene details. This methodology enables the generation of temporally coherent and contextually enriched video scenes, aligning closely with the intended actions described in the input text. ART•V~\cite{weng2023art} focuses on an innovative approach that addresses the challenges of modeling complex long-range motions by sequentially generating video frames, each conditioned on its predecessors. This strategy ensures the production of continuous and simple motions, maintaining coherence between adjacent frames. DynamiCrafter~\cite{xing2023dynamicrafter} focuses on a dual-stream image injection mechanism that integrates both a text-aligned context representation and visual detail guidance. This approach ensures that the animated content remains visually coherent with the input image while being dynamically consistent with the textual description. 

% motion generation improvement 
Several works have concentrated on improving motion generation in T2V. PixelDance~\cite{zeng2023make} focuses on enriching video dynamics by integrating image instructions for the video's first and last frames alongside text instructions. This method allows the model to capture complex scene transitions and actions, enhancing the motion richness and temporal coherence in the generated videos. MoVideo~\cite{liang2023movideo} using depth and optical flow information derived from a key frame to guide the video generation process. Initially, an image is generated from the text, and then depth and optical flows are extracted from this image. MicroCinema~\cite{wang2023microcinema} employs a two-stage process, initially generating a key image from text, then using both the image and text to guide the video creation, focusing on capturing motion dynamics. ConditionVideo~\cite{peng2023conditionvideo} separates video motion into background and foreground components, enhancing clarity and control over the generated video content. DreamVideo~\cite{wei2023dreamvideo} decouples the video generation task into subject learning and motion learning. The motion learning aspect is specifically targeted to adapt the model to new motion patterns effectively. They introduce a motion adapter, which, when combined with appearance guidance, enables the model to concentrate solely on learning the motion without being influenced by the subject's appearance. TF-T2V~\cite{wang2023recipe} enables the learning of intricate motion dynamics without relying on textual annotations, employing an image-conditioned model to capture diverse motion patterns effectively, which includes a temporal coherence loss that strengthens the continuity across frames. GPT4Motion~\cite{lv2023gpt4motion} employs GPT-4 to generate Blender scripts, which are then used to drive Blender's physics engine, simulating realistic physical scenes that correspond to the given textual descriptions. The integration of these scripts with Blender's simulation capabilities ensures the production of videos that are not only visually coherent with the text prompts but also adhere to physical realism. However, challenges remain, particularly in human motion generation. Text2Performer~\cite{jiang2023text2performer} diverges from traditional discrete VQ-based models by generating continuous pose embeddings, enhancing motion realism and temporal coherence in the generated videos. MotionDiffuse~\cite{zhang2024motiondiffuse} focuses on a probabilistic strategy that facilitates the creation of varied and lifelike motion sequences from text descriptions. Integrating a Denoising Diffusion Probabilistic Model (DDPM) with a cross-modality transformer architecture allows the system to produce intricate, ongoing motion sequences aligned with textual inputs, ensuring both high fidelity and precise controllability in the resulting animations. 

\subsubsection{Complex Scene}
Generating videos with complex scenes is exceptionally challenging due to the intricate interplay of elements, requiring sophisticated understanding and replication of detailed environments, dynamic interactions, and variable lighting conditions with high fidelity. In the initial phases of research, authors in \cite{vondrick2016generating} proposed a GAN network that incorporates a novel spatio-temporal convolutional architecture. This architecture effectively captures the dynamics of both the foreground and background elements in the scene. By training the model on large datasets of unlabeled video, it learns to predict plausible future frames from static images, creating videos with realistic scene dynamics. This method allows the model to handle complex scenes by understanding and generating the temporal evolution of different components within a scene.

Subsequently, as LLMs evolved, researchers delved into leveraging their capabilities to enhance generative models in producing complex scenes. VideoDirectorGPT~\cite{lin2023videodirectorgpt} leverages LLM for video content planning, producing detailed scene descriptions, and entities with their layouts, and ensuring visual consistency across scenes. By employing a novel Layout2Vid generation technique to ensure spatial and temporal consistency across scenes it produces rich, narrative-driven video content. Similarly, FlowZero~\cite{lu2023flowzero} enhances alignment between spatio-temporal layouts and text prompts through a novel framework for zero-shot T2V synthesis. Integrating LLM and image diffusion models, FlowZero initially translates textual prompts into detailed Dynamic Scene Syntax (DSS), outlining scene descriptions, object layouts, and motion patterns. It then iteratively refines these layouts in line with the text prompts via self-refinement processes, thereby synthesizing temporally coherent videos with intricate motions and transformations.

VideoDrafter~\cite{long2024videodrafter} converts input prompts into comprehensive scripts by utilizing LLM, identifies common entities, and generates reference images for each entity. VideoDrafter then produces videos by considering the reference images, descriptive prompts, and camera movements through a diffusion process, ensuring visual consistency across scenes. SceneScape~\cite{fridman2024scenescape} emphasizes the generation of videos for more complex scenarios within 3D scene synthesis. Using pre-trained text-to-image and depth prediction models ensures the generation of videos that maintain 3D consistency via a test-time optimization process. It employs a progressive strategy where a unified mesh representation of the scene is continuously built and updated with each frame, thereby guaranteeing geometric plausibility.
% Video
% 1. FlowZero: Zero-Shot Text-to-Video Synthesis with LLM-Driven Dynamic Scene Syntax  complex spatiotemporal
% 2. VideoDirectorGPT: Consistent Multi-scene Video Generation via LLM-Guided Planning 一致多场景， 长视频生成
% 3. VideoDrafter: Content-Consistent Multi-Scene Video Generation with LLM . 统一角色多场景切换， 也是多个背景， 解决外观一致性
% 4. SceneScape: Text-Driven Consistent Scene Generation 。主要是连续的3D场景生成
\subsubsection{Multiple Objects}
% T2I
% 1.Detector Guidance for Multi-Object Text-to-Image Generation
% 
% Non text-condition
% 1.Multi-object Video Generation from Single Frame Layouts
% 2.Xp-GAN: Unsupervised Multi-object Controllable Video Generation 主要是控制不是生成
% 
% Video
% 1. UniVG: Towards UNIfied-modal Video Generation Multi-modal inputs
% 2. VideoDreamer: Customized Multi-Subject Text-to-Video Generation with Disen-Mix Finetuning  multi-subjects
As for every frame of a video, generating multiple objects presents several challenges such as attribute mixing, object mixing, and object disappearance. Attribute mixing occurs when objects inadvertently and mistakenly adopt characteristics from other objects. Object mixing and disappearance involve the blending of distinct objects, which results in the creation of peculiar hybrids and inaccurate object counts. To solve these problems, Detector Guidance (DG)~\cite{liu2023detector} integrates a latent object detection model to enhance the separation and clarity of different objects within generated images. Their approach manipulates cross-attention maps to refine object representation, demonstrating significant improvements in generating distinct objects without human intervention. 

The complexity of video synthesis necessitates capturing the dynamic spatio-temporal relationships between objects. MOVGAN~\cite{wu2023multi}, drawing inspiration from layout-to-image generation advancements, innovates by employing implicit neural representations alongside a self-inferred layout motion technique. This approach enables the generation of videos that not only depicts individual objects but also accurately represents their interactions and movements over time, enhancing the realism and depth of the synthesized video content.

In the realm of T2V generation, addressing the visual features of multiple subjects within a single video frame is crucial due to the attribute binding problem. VideoDreamer~\cite{chen2023videodreamer} leverages Stable Diffusion with latent-code motion dynamics and temporal cross-frame attention, further customized through Disen-Mix Finetuning and optional Human-in-the-Loop Re-finetuning strategies. It successfully generates high-resolution videos by maintaining temporal consistency and subject identity without artifacts.

UniVG~\cite{ruan2024univg} tackles multi-object generation challenges by enhancing its base model with Multi-condition Cross-Attention for tasks requiring high freedom, enabling effective management of complex scenarios involving multiple objects derived from text and image inputs. For tasks with lower freedom, it introduces Biased Gaussian Noise to more effectively preserve content, aiding in tasks such as image animation and super-resolution. These innovations allow UniVG to produce semantically aligned, high-quality videos across a variety of generative tasks, effectively addressing the intricacies of multi-object scenarios.

% \subsubsection{Physical World Simulation and Causal Understanding}
\subsubsection{Rational Layout}
% Placeholder
% 1. LLM-grounded Video Diffusion Models 符合逻辑的layout生成 。在生成视频
% 
% understanding
% 1. StyleCrafter: Enhancing Stylized Text-to-Video Generation with Style Adapter  Prompt Understanding Optimization
% 2. FlowZero: Zero-Shot Text-to-Video Synthesis with LLM-Driven Dynamic Scene Syntax  增强理解 complex spatiotemporal
% 3. Make-Your-Video: Customized Video Generation Using Textual and Structural Guidance 生成视频时增强对用户输入的理解
% 4. CogVideo: Large-scale Pretraining for Text-to-Video Generation via Transformers 语义生成理解，  比如复杂运动，  语义视频alignment
% 5. Imagine this! scripts to compositions to videos 主要semantic信息等等， 传统方法 （其实并不直接属于这个类别）

Ensuring high-quality video output in T2V conversion hinges on the ability to generate rational layouts based on textual instructions. Craft~\cite{gupta2018imagine} is designed for generating videos from textual descriptions, learning from video-caption data to predict the temporal layout of entities within a scene, retrieve spatio-temporal segments from a video database, and fuse them to generate scene videos. It incorporates the Layout Composer, a model that generates plausible scene layouts by understanding spatial relationships among entities. Utilizing a sequential approach that combines text embeddings and scene context, Craft accurately predicts the locations and scales of characters and objects, facilitating the creation of visually coherent and contextually accurate videos.

FlowZero~\cite{lu2023flowzero} generate layouts using LLMs to transform textual prompts into structured syntaxes for guiding the generation of temporally-coherent videos. This process includes frame-by-frame scene descriptions, foreground object layouts, and background motion patterns. Specifically, for foreground layouts, LLMs generate a sequence of frame-specific layouts that outline the spatial arrangement of foreground entities in each frame. These layouts consist of bounding boxes defining the position and size of the prompt-referenced objects. This structured approach ensures that foreground objects adhere to the visual and spatio-temporal cues provided in the text, enhancing the video's coherence and fidelity.

Nonetheless, authors in~\cite{lian2023llm} highlight that existing models face challenges with sophisticated spatiotemporal prompts, frequently resulting in limited or erroneous movements, for instance, their inability to accurately represent objects transitioning from left to right. Addressing these shortcomings, they propose LLM-grounded Video Diffusion (LVD), a novel method that enhances neural video generation from text prompts by first generating dynamic scene layouts (DSLs) with a Large Language Model (LLM), then using these DSLs to guide a diffusion model for video generation. This approach addresses the limitations of current models in generating videos with complex spatiotemporal dynamics and achieves significantly better performance in generating videos that closely align with the desired attributes and motion patterns.

\begin{table*}[!t]
    \centering
    \caption{Comparison of existing text-to-video datasets.}
    \begin{tabular}{r r r r r r r r r r r r r}
    \hline
           Dataset  & Domain & Annotated &\#Clips & \#Sent & $\mathrm{Len_{C}}$(s)& $\mathrm{Len_{S}}$ & \#Videos& Resolution & FPS & Dur(h) & Year & Source\\
    \hline
         CV-Text~\cite{yu2023celebv}  & Face & Generated & 70K & 1400K& - & 67.2 & - & 480P & - & - & 2023 & Online\\
         \hdashline
         
MSR-VTT~\cite{xu2016msr}  & Open & Manual & 10K & 200K & 15.0s & 9.3 & 7.2K & 240P  & 30 & 40 & 2016 & YouTube\\ 
        DideMo~\cite{anne2017localizing} & Open & Manual & 27K & 41K & 6.9s & 8.0 & 10.5K &-& & 87 & 2017 & Flickr \\
         Y-T-180M~\cite{zellers2021merlot}  & Open & ASR & 180M & - & - & - & 6M & - & - & - & 2021 & YouTube\\
         WVid2M~\cite{bain2021frozen} & Open & Alt-text & 2.5M & 2.5M & 18.0 & 12.0 & 2.5M & 360P & - & 13K & 2021 & Web\\
         H-100M~\cite{xue2022advancing}  & Open & ASR & 103M & - & 13.4 & 32.5 & 3.3M & 720P & - & 371.5K & 2022 &YouTube\\
         InternVid~\cite{wang2023internvid} & Open & Generated & 234M& -&11.7 &17.6 & 7.1M & *720P & - & 760.3K& 2023& YouTube \\
         H-130M~\cite{wang2023videofactory} & Open & Generated & 130M & 130M & - & 10.0 & - & 720P & - & - & 2023& YouTube \\
         Y-mP~\cite{xu2023youku}  & Open & Manual & 10M & 10M & 54.2 & - & - & - & - & 150K & 2023& Youku\\
         V-27M~\cite{chen2024vast}  & Open & Generated & 27M & 135M & 12.5 & - & - & - & - & - &2024& YouTube\\
         P-70M~\cite{chen2024panda} & Open & Generated & - & 70.8M & 8.5 & 13.2 & 70.8M & 720P & - & 166.8K & 2024 & YouTube \\
         \hdashline
         LSMDC~\cite{rohrbach2017movie}  & Movie & Manual & 118K & 118K & 4.8s & 7.0 & 200 & 1080P & - & 158 & 2017 & Movie\\

         MAD~\cite{soldan2022mad} & Movie & Manual & - & 384K & - & 12.7 & 650 & - & - & 1.2K &2022 & Movie\\
         \hdashline
        UCF-101~\cite{soomro2012ucf101} & Action& Manual & 13K & - & 7.2s & - & - &240P & 25& 27 &  2012  & YouTube\\
        ANet-200~\cite{caba2015activitynet}  & Action & Manual & 100K & - & - & 13.5 & 2K &{*}720P &30 & 849 & 2015& YouTube\\
    
         Charades~\cite{sigurdsson2016hollywood} & Action & Manual & 10K & 16K & - & - &10K & - & - & 82 & 2016& Home\\
         Kinetics~\cite{kay2017kinetics}  & Action & Manual & 306K & - & 10.0s & - & 306K & - & - & - & 2017& YouTube \\
       
         ActNet~\cite{krishna2017dense} & Action & Manual & 100K & 100K & 36.0s & 13.5 & 20K & - & - & 849 & 2017 & YouTube\\
         C-Ego~\cite{sigurdsson2018charades} & Action & Manual & - & - & - & - &8K & 240P & - & 69 & 2018 & Home\\
        SS-V2~\cite{goyal2017something}  & Action & Manual & - & - & - & - &220.1K & - & 12 & - & 2018& Daily\\
        \hdashline

         How2~\cite{sanabria2018how2}  & Instruct & Manual & 80K & 80K & 90.0 & 20.0 & 13.1K & - & - & 2000 & 2018& YouTube\\

         HT100M~\cite{miech2019howto100m} & Instruct & ASR & 136M & 136M & 3.6 & 4.0 & 1.2M & 240P & - & 134.5K & 2019& YouTube \\
         \hdashline

         YCook2~\cite{zhou2018towards}  & Cooking & Manual & 14K & 14K & 19.6 & 8.8 & 2K & - & - & 176 & 2018& YouTube\\
         E-Kit~\cite{damen2018scaling}  & Cooking & Manual & 40K & 40K & - & - & 432 & *1080P& 60 & 55 & 2018 & Home\\
         \hline
         
    \end{tabular}
    \label{tab:dataset_comparasion}
\end{table*}

\subsection{Datasets and Metrics} 
\label{sec:dataset-metrics}
\subsubsection{Datasets}
We comprehensively review the T2V datasets and mainly categorize them into six streams based on the collected domain: \emph{Face}, \emph{Open}, \emph{Movie}, \emph{Action}, \emph{Instruct}, and \emph{Cooking}.
Table~\ref{tab:dataset_comparasion} summarizes datasets in 12 dimensions, and the details of each dataset are described as follows:

{\textbf{CV-Text~\cite{yu2023celebv}}} is a high-quality dataset of facial text-video pairs, comprising 70,000 in-the-wild face video clips with a resolution of at least 512$\times$512.
Each clip is paired with 20 generated descriptions whose average length is around 67.2.

{\textbf{MSR-VTT~\cite{xu2016msr}} provides 10K web video clips with 40 hours and 200K clip-sentence pairs in total.
Each clip is accompanied by approximately 20 natural sentences for description.
The number of videos is around 7.2K, and the average of each clip and sentence is 15.0s and 9.3 words respectively.  

{\textbf{DideMo~\cite{anne2017localizing}}} consists of over 10,000 unedited, personal videos in diverse visual settings with pairs of localized video segments and referring expressions.

{\textbf{YT-Tem-180M~\cite{zellers2021merlot}}} was collected from 6 million public YouTube videos, 180 million clips, and annotated by ASR.

{\textbf{WebVid2M~\cite{bain2021frozen}}} consists of 2.5M video-text pairs.
The average length of each video and sentence is 18.0s and 12.0 words respectively.
The raw descriptions of each video are collected from the Alt-text HTML attribute associated with web images.

{\textbf{HD-VILA-100M}~\cite{xue2022advancing}} is a large-scale text-video dataset collected from YouTube consisting of 100M high-resolution (720P) video clips and sentence pairs from 3.3 million videos with 371.5K hours and 15 popular categories in total.

{\textbf{InternVid~\cite{wang2023internvid}}} is a large-scale video-centric multimodal dataset that enables learning powerful and transferable video-text representations for multimodal understanding and generation. 
InterVid contains over 7 million videos lasting nearly 760K hours, yielding 234M video clips accompanied
by detailed descriptions of a total of 4.1B words.

{\textbf{HD-VG-130M~\cite{wang2023videofactory}}} comprises 130 million text-video pairs from the open domain with a high resolution (1376$\times$768).
Most descriptions in the dataset are around 10 words.

{\textbf{Youku-mPLUG~\cite{xu2023youku}}} is the first released Chinese video-language pretraining dataset collected from Youku~\cite{youku}.
It contains 10 million Chinese video-text pairs filtered from 400 million raw videos across a wide range of 45 diverse categories.
The average time of each video is around 54.2s.

{\textbf{VAST-27M~\cite{chen2024vast}}} consists of a total of 27M video clips covering diverse categories, and each clip paired with 11 captions (including 5 vision, 5 audio, and 1 omni-modality caption).
The average lengths of vision, audio, and omni-modality captions are 12.5, 7.2, and 32.4 respectively.

{\textbf{Panda-70M~\cite{chen2024panda}}} is a high-quality video-text dataset originally curated from HD-VILA-100M.
It consists of a total duration of 166.8Khr of 70.8M videos with paired high-quality text captions.
The time of each video is 8.5s and the length of each sentence is 13.2 words on average.

{\textbf{LSMDC~\cite{rohrbach2017movie}}} consists of around 118K video clips aligned to sentences sourced from 200 movies.
The total time of videos is around 158 hours, and the average of each clip and sentence is 4.8s and 7.0 words respectively.

{\textbf{MAD~\cite{soldan2022mad}}} is derived from movies, and contains more than 384K sentences anchored on more than 1.2K hours of 650 videos.
The length of each sentence is 13.2 words on average.

{\textbf{UCF-101~\cite{soomro2012ucf101}}} is a dataset of human actions, collected samples from YouTube~\cite{youtube} that encompass 101 action categories, including human sports, musical instrument playing, and interactive actions.
It consists of over 13K clips and 27 hours of video data. 
The resolution and frame rate of UCF-101 is 320$\times$240 and 25 Frame-Per-Second (FPS).

{\textbf{ActNet-200~\cite{caba2015activitynet}}} provides a total of 849 hours of video, where 68.8 hours of video contain 203 human-centric activities. 
Around 50\% of the videos are in HD resolution (1280$\times$720), while the majority have a frame rate of 30 FPS.

{\textbf{Charades~\cite{sigurdsson2016hollywood}}} is collected from household activities of 267 people, which consists of around 10K videos covering 157 daily actions with an average length of 30.1s. 

{\textbf{Kinetics~\cite{kay2017kinetics}}} contains 400 human action classes, with at least 400 video clips for each action.
Each clip lasts around 10s and is taken from a different YouTube video.

{\textbf{SS-V2~\cite{goyal2017something}}} is a large collection of labeled video clips that show humans performing pre-defined basic 174 actions with everyday objects. 
The dataset was created by a large number of crowd workers, containing around 220.1K videos.

{\textbf{ActivityNet~\cite{krishna2017dense}}} contains 20K videos amounting to 849 video hours with 100k total descriptions, each with its unique start and end time.
For each sentence, the length is 13.5 words and for descriptions, is 36s on average.

{\textbf{Charades-Ego~\cite{sigurdsson2018charades}}} consists of 8K videos (4K pairs of third and first-person videos) total.
In these videos, with over 364 pairs involving more than one person in the video.
The average time of each video is around 31.2s.

{\textbf{How2~\cite{sanabria2018how2}}} covers a wide variety of instructional topics.
It consists of 80K clips with 2K hours in total, and the length of each video is around 90s on average.

{\textbf{HowTo100M~\cite{miech2019howto100m}} consists of 136 million video clips collected from 1.22M narrated instructional web videos depicting humans performing and describing over 23K different visual tasks. 
Each clip is paired with a text annotation in the form of automatic speech recognition (ASR).

{\textbf{YouCook2}~\cite{zhou2018towards}} contains 2000 videos that are nearly equally distributed over 89 recipes with a total length of 176 hours.
The recipes are from four major cuisines (\emph{i.e.}, African, American, Asian and European)
and have a large variety of cooking styles, methods, ingredients, and cookware. 

{\textbf{Epic-Kitchens~\cite{damen2018scaling}}} is an egocentric video recorded by 32 participants in the kitchen environments.
It contains 55 hours of video and a total of around 40.0K action segments.
Major videos were recorded in a full HD resolution of 1920$\times$1080.

\subsubsection{Metrics} 
The metrics used to evaluate the quality of video generated from T2V models can be mainly categorized into two aspects: quantitative and qualitative.
In the former, the evaluators are typically presented with two or more generated videos to compare against videos synthesized by other competitive models.
Observers generally engage in voting-based assessments regarding the realism, natural coherence, and text alignment of the videos.
Although human evaluation has been utilized in various works~\cite{singer2022make,wang2023videofactory, xing2023simda, wu2023tune}, the time-consuming and labor-intensive aspects restrict its wide application.
Besides, qualitative measurements fail to give a comprehensive evaluation of the model at times~\cite{xing2023survey}.
Therefore, we restrict our review to the quantitative metrics.

These reviewed metrics are categorized into \emph{image-level} and \emph{video-level}.
Generally, the former is utilized to evaluate generated videos frame-wise, and the latter focuses on the comprehensive evaluation
of the video quality:

\textbf{Peak Signal-to-Noise Ration (PSNR)~\cite{wang2004image}.} 
Commonly known as PSNR, was designed to quantify reconstruction quality for the frame of generated videos.
Given an original image $I_{o}$ with $M\times N$ pixels and the generated image $I_{g}$, the PSNR can be calculated as: 
\begin{equation}
    \mathbf{PSNR} = 10\cdot \log_{10}(\frac{\mathrm{MAX}_{I_{o}}^{2}}{\mathrm{MSE}}),
\end{equation}
where $\mathrm{MSE} = \frac{1}{MN}\sum_i\sum_j[I_{o}(i,j) - I_{g}(i,j)]^2$.
$\mathrm{MAX}_{I_{o}}$ is the possible maximal value of the original image. 
$I_{o}(i, j)$ and $I_{g}(i, j)$ denote the pixel value on the position $(i, j)$.

\textbf{Structural Similarity Index (SSIM)~\cite{wang2004image}.}
SSIM is usually utilized to measure the similarity between two images, and also as a method to quantify the quality from a perceptive view.
Specifically, the definition of SSIM is:
\begin{equation}
    \mathbf{SSIM} = [L(I_o, I_g)]^\alpha \cdot [C(I_o, I_g)]^\beta \cdot [S(I_o, I_g)]^\gamma,
\end{equation}
where $L(I_o, I_g)$ is utilized to compare the luminance, $C(I_o, I_g)$ reflects the contrast difference, $S(I_o, I_g)$ measures the structure similarity between the original image $I_{o}$ and the generated image $I_{g}$.
Practically, the parameters are set as $\alpha=1$, $\beta=1$, and $\gamma=1$.
Then, the SSIM can be simply represented as:
\begin{equation}
    \mathbf{SSIM} = \frac{(2\Bar{{I_{o}}}\Bar{{I_{g}}}+\delta_{1})(2\Sigma_{I_{o}I_{g}} + \delta_{2})}{(\Bar{I}_{o}^{2}+\Bar{I}_{g}^{2} + \delta_{1})(\Sigma_{I_{o}}^{2}+\Sigma_{{I}_{g}}^{2} + \delta_{2})},
\end{equation}
where $\Bar{I}$ denotes the average value of the image $I$. 
$\Sigma_{I_{o}}^{2}$ and $\Sigma_{I_{g}}^{2}$ are the variance of $I_{o}$ and $I_{g}$ respectively. 
We use $\Sigma_{I_{o}I_{g}}$ to denote the covariance.
Lastly, $L$ is the maximum intensity value in $\delta_{i} = (K_{i}L)^2$ and $K_{i} \ll 1$.

\textbf{Inception Score (IS)~\cite{salimans2016improved}.} IS is developed to measure both image-generated quality and diversity.
Specifically, the pre-trained Inception network~\cite{szegedy2016rethinking} is applied to get the conditional label distribution $p(y|I_{g})$ of each generated image.
IS can be formulated as:
\begin{equation}
    \mathbf{IS}=\exp(\mathbb{E}_{I_{g}}[\mathrm{KL}(p(y|I_{g})||p(y)]),
\end{equation}
where $\mathrm{KL}$ denotes the KL-divergence~\cite{hershey2007approximating}. 

\textbf{Fréchet Inception Distance (FID)~\cite{heusel2017gans}.}
Compared with IS, FID provides a more comprehensive and accurate evaluation as it directly considers the similarity between the generated images and original images.

\begin{equation}
    \mathbf{FID} = ||\Bar{I}_{o} - \Bar{I}_{g}||^2 + \mathrm{Tr}(\Sigma_{{I}_{g}} + \Sigma_{I_{r}} - 2(\Sigma_{I_{g}}\Sigma_{I_{r}})^{1/2}).
\end{equation}
Here, $\mathrm{Tr}$ is the trace of a matrix.

\textbf{CLIP Score~\cite{radford2021learning}.} CLIP Score has been widely used to measure the alignment between images and sentences.
Based on the pre-trained CLIP embedding, it can be calculated as:
\begin{equation}
    \mathbf{CLIP}_{score} = \mathbb{E}[\max(\cos(\mathcal{E}_{I}, \mathcal{E}_{S}), 0)],
\end{equation}
where $\mathcal{E}_{I}$ and $\mathcal{E}_{S}$ denote the embedded features from the image and the sentence respectively, $\cos({\mathcal{E}_{I}, \mathcal{E}_{S}})$ is the cosine similarity.

\textbf{Video Inception Score (Video IS)~\cite{saito2020train}.} 
Typically, Video IS calculates the IS of the generated videos based on the features extracted from C3D~\cite{tran2015learning}.

\textbf{{Fréchet Video Distance (FVD).}}
Based on the extracted features from pre-trained Inflated-3D Convnets (I3D)~\cite{carreira2017quo}  Fréchet Video Distance (FVD)~\cite{unterthiner2019fvd} scores can be computed by combining the means and covariance matrices:
\begin{equation}
    \mathbf{FVD} = ||\Bar{\mathcal{V}} - \Bar{\mathcal{V}}^{*}||^2 + \mathrm{Tr}(\Sigma_{\mathcal{V}} + \Sigma_{\mathcal{V}^{*}} - 2(\Sigma_{\mathcal{V}}\Sigma_{\mathcal{V}^{*}})^{1/2}). 
\end{equation}
Here, we use the $\Bar{\mathcal{V}}$ and $\Bar{\mathcal{V}}^{*}$ to denote the exception of the realistic videos $\mathcal{V}$ and the generated videos $\mathcal{V}^{*}$ respectively.

\textbf{{Kernel Video Distance (KVD)~\cite{unterthiner2018towards}.}}
KVD adopts the kernel method to evaluate the performance of generated model.
Given the kernel function $\Phi$, and sets $\{\mathcal{F}, \mathcal{F}^{*}\}$ that are features extracted from the realistic videos and generated videos via the pre-trained I3D and combining   Maximum Mean Discrepancy (MMD), KVD can be formulated as:
\begin{equation}
    \mathbb{E}_{f,f'}[\Phi(f, f')] + \mathbb{E}_{f^{*}, f^{*}{'}}[\Phi(f^{*}, f^{*}{'})] - 2\mathbb{E}_{f, f^{*}}[\Phi(f, f^{*})].
\end{equation}
Concretely, $f$ and $f^{*}$ are sampled from $\mathcal{F}$ and $\mathcal{F}^{*}$ respectively.

\textbf{Frame Consis Score (FCS)~\cite{wu2023tune}.}
FCS calculates the cosine similarity of the CLIP image embeddings for all pairs of video frames to measure the consistency of edited videos.

\section{Challenges and Open Problems}
\label{sec:chall_pro}
At beginning of this section, we will generally go through all existing common problems still not being addressed, even for SOTA work Sora.
% 可以小点堆砌。  参考https://arxiv.org/pdf/2402.03578.pdf

\subsection{Unsolved Problems from Sora}
\begin{figure*}[ht!]
    \centering
    \includegraphics[width=\textwidth]{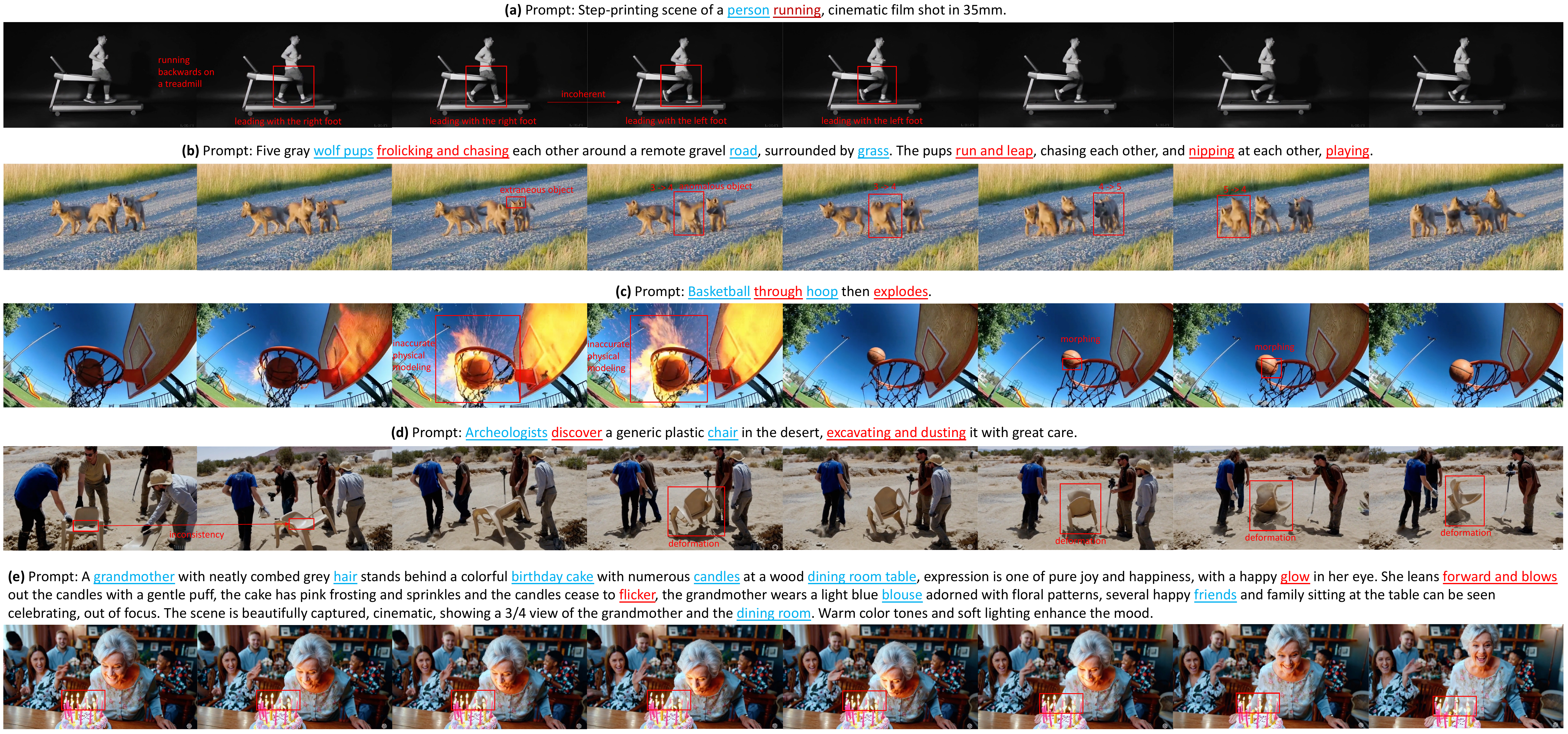}
    \caption{Screenshots of Sora generated video with its prompts from~\cite{sora_web}}
    \label{fig:sora_weakness}
\end{figure*}

In Fig~\ref{fig:sora_weakness}, there are five weaknesses based on the video demonstrated by OpenAI on their website ~\cite{sora_web}. 

{\textbf{Unrealistic and incoherent motion:}} In Fig~\ref{fig:sora_weakness}(a), we observe a striking example of unrealistic motion: a person appears to be running backwards on a treadmill but paradoxically, the running action is forward, a physically implausible scenario. Typically, treadmills are designed for forward-facing use, as depicted on the left side of the figure. This discrepancy highlights a prevalent issue in T2V synthesis, where current LLMs excel at understanding and explaining the physical laws of motion but struggle to accurately render these laws in visual or video formats.

Additionally, the video exhibits incoherent motions; for instance, a person’s running pattern should display a consistent sequence of leg movements. Instead, there’s an abrupt shift in the positioning of the legs, disrupting the natural flow of motion. Such frame-to-frame inconsistencies underscore another significant challenge in T2V conversion: maintaining motion coherence throughout the video sequence.

{\textbf{Intermittent object appearances and disappearances:}} In Fig~\ref{fig:sora_weakness}(b), we see a multi-object scene characterized by intermittent appearances and disappearances of objects, which detracts from the generation accuracy of the video from text. Despite the prompt specifying ``five wolf pups" initially, only three are visible. Subsequently, an anomaly occurs where one wolf inexplicably exhibits two pairs of ears. Progressing through the frames, a new wolf unexpectedly appears in front of the middle wolf, followed by another appearing in front of the rightmost wolf. Ultimately, the first wolf, which is displayed from the middle of the screen, disappears from the scene.

Such spontaneous manifestations of animals or people, particularly in scenes with numerous entities, pose a significant challenge. When a video necessitates a precise count of objects or characters, the occurrence of such unplanned elements can disrupt the narrative, leading to outputs that are not only inaccurate but also inconsistent with the specified prompt. The introduction of unexpected objects, especially if they contradict the intended storyline or content, risks misrepresenting the intended message, undermining the video’s integrity and coherence.

{\textbf{Unrealistic phenomena:}} Fig~\ref{fig:sora_weakness}(c) showcases two sequences of snapshot frames, illustrating the issues of inaccurate physical modeling and unnatural object morphing. Initially, the first sequence of four frames depicts a basketball passing through the hoop and igniting into flames, as specified by the prompt. However, contrary to the expected explosive interaction, the basketball passes through the hoop unscathed. Subsequently, the next set of four frames reveals another basketball traversing the hoop, but this time it directly passes through the rim which unexpectedly morphs during the sequence. Additionally, in this snapshot, the basketball fails to explode as dictated by the prompt.

This scenario exemplifies the challenges of inaccurate physical modeling and unnatural object morphing. In the depicted case, Sora seems to induce some unnatural changes to objects, which can significantly detract from the realism of the videos. 

% Such inaccuracies are particularly problematic when the generated videos are intended for use in marketing, where authenticity and adherence to the specified visual effects are crucial. 改一下例子

{\textbf{Limited understanding of objects and characteristics:}} Fig~\ref{fig:sora_weakness}(d) illustrates the limitations of Sora in accurately comprehending objects and their inherent characteristics, with a focus on texture. The sequence reveals a plastic chair that initially appears stable but then undergoes bending and shows inconsistent shapes between the initial frames, alongside an unnatural floating without any visible support. Subsequently, the chair is depicted undergoing continuous, extreme bending.

This represents a failure to correctly model the chair as a rigid, stable object, leading to implausible physical interactions. Such inaccuracies can result in videos that appear surreal and are therefore not suitable for practical use. Although minor flaws might be corrected with additional design tools, significant errors, particularly with prominent objects displaying unrealistic behavior across multiple frames, could render the video ineffective. Consequently, achieving the desired outcome might require numerous iterations to correct these pronounced inconsistencies

{\textbf{Incorrect interactions between multi-objects:}} 
Fig~\ref{fig:sora_weakness}(d) illustrates the model's inability to accurately simulate complex interactions involving multiple objects. The sequence intended to show a `grandmother' character blowing out candles. Ideally, the candle flames should react to the airflow, either by flickering or being extinguished. However, the video fails to depict any interaction with the external environment, displaying the flames as unnaturally static throughout the scene.

This highlights Sora's challenges in rendering realistic interactions, particularly in scenes with multiple active elements or complex dynamics. The difficulty is amplified in scenarios involving numerous moving subjects, intricate backgrounds, or interactions involving various textures. Such shortcomings can lead to unrealistic and sometimes unintentionally humorous outcomes, undermining the effectiveness of the video, especially in contexts requiring a high degree of realism or accurate representation of physical interactions.

\subsection{Data access privacy}
% 1.
% Privacy-preserving Collaborative T2V Generation
Inspired by advancements in large language models (LLMs), Sora is trained using internet-scale datasets~\cite{videoworldsimulators2024}. However, the vast expanse of the Internet's public data constitutes only a fraction of the total available information; the bulk is private data from individuals, companies, and institutions. Compared to public datasets, private ones are richer in diversity and contain fewer duplicates. However, they also harbor a significant amount of sensitive personal information, particularly in image and video formats, which tends to have more personalized content than plain text. This is a critical distinction, as the latter is predominantly used for training LLMs. Therefore, when the public data resources are fully utilized and there is an aim to further enhance model capabilities, especially in terms of generalizability, it becomes crucial to devise strategies for leveraging private, non-sensitive data in a manner that rigorously protects privacy.

Federated Learning (FL), as introduced in \cite{mcmahan2017communication}, offers a promising solution that enables thousands of clients to collaboratively contribute to a global model using their own data, without the need to transfer any raw data at any stage. Recent studies have validated the efficacy of FL in fine-tuning LLMs ~\cite{zhang2023towards} and enhancing diffusion models~\cite{huang2024federated}. While FL can effectively address the challenges of distributed private data accessibility, it still faces significant problems, including network bottlenecks, heterogeneous data, the intermittent availability of devices and so on.

% \subsection{Text-to-Video-to-Action}
% 网站上 sora。Simulating digital worlds

% \subsubsection{Physical World Simulation and Causal Understanding}

\subsection{Simultaneous Multi-shot Video Generation}
Multi-shot video generation stands as a significant challenge within the T2V sector, a realm where video generation itself remains largely uncharted. This scenario persisted until Sora showcased its proficiency in this domain. With its advanced linguistic comprehension, Sora is capable of generating videos that incorporate multiple shots, consistently maintaining the characters and visual style throughout the sequence. Such capability signifies a significant stride in video generation, offering a coherent and continuous visual narrative.

Despite these advancements, Sora has yet to master the creation of simultaneous multi-shot videos featuring identical characters and consistent visual styles. This capability is crucial for specific fields, such as robotics, where learning from demonstration is essential, and in autonomous vehicle simulations, where consistent visual representation plays a pivotal role in the system's effectiveness and reliability. The development of this feature would mark a substantial breakthrough, broadening the applicability of T2V technologies in critical, real-world applications.

% 首先，目前就 Sora 所呈现的，虽然有多机位效果，但都是单一情节单一镜头。

% 而《三体》有多人视角，叙事复杂，靠 AI 生成一条龙搞定并不现实。

% 论文
% 1. Stable Video Diffusion: Scaling Latent Video Diffusion Models to Large Datasets

\subsection{Multi-Agent Co-creation}
Agents powered by LLMs can perform tasks independently based on human knowledge, while multi-agent systems bring together individual LLM agents, utilizing their collective capabilities and specialized skills through collaboration~\cite{hong2023metagpt, li2023camel, talebirad2023multi}. In multi-agent systems, collaboration and coordination are pivotal, enabling a collective of agents to undertake tasks that surpass the capabilities of any single entity, and agents are endowed with unique abilities and assume specific roles, working collaboratively to achieve shared goals~\cite{han2024llm}. The efficacy of multi-agent systems in reflecting and augmenting real-world tasks requiring collaborative problem-solving or deliberation has been substantiated across various domains. Notably, this includes areas like software development~\cite{hong2023metagpt, li2023camel} and negotiation games~\cite{abdelnabi2023llm}. 

Despite its potential, the application of multi-agent systems in the T2V generation field remains largely unexplored due to unique challenges. Take filmmaking as an instance: agents assume distinct roles, such as directors, screenwriters, and actors. Each actor must generate their segment in accordance with directives from the director, subsequently integrating these segments into a cohesive video. The primary challenge lies in ensuring that each actor agent not only preserves frame-to-frame coherence within its own output but also aligns with the collective vision to maintain global video style consistency, logical screen layout, and uniformity. This is particularly difficult as outputs from generative models can vary significantly, even when prompted similarly.

% Complex Interactions Between Objects: Generating images or videos with multiple objects requires understanding and rendering the complex interactions and spatial relationships between these objects. Each object may have its own attributes and context, which must be accurately represented in relation to other objects within the scene.
% 动作生成一致性以及协调，导演需要指导每个人的单独角色的渲染， 然后每个子视频需要进行拼接成一个完整的视频，不光要视频帧之间的连贯性，还需要和其他子视频配合，这个是个大挑战, 风格一致统一控制
% 现在都是对输入的text 直接进行 prompt 拆解或者扩展，但是需要agent扩展
% CoT

\section{Future Directions}
\label{sec:future}
% \noindent{\textbf{Foundation Models Cascaded Bias}}
% 1. Foundation models pipeline

% \noindent{\textbf{High-quality Video Generation with Long Prompts}}

\subsection{Robot Learning from Visual Assistance}
% 1. multi-view shots
% https://arxiv.org/pdf/2402.07127.pdf
% 先说demo 机器人方向
% 这章需要一个图
Traditional methods of programming robots for new tasks demand extensive coding expertise and a significant investment of time~\cite{ravichandar2020recent}. These approaches require users to meticulously define each step or movement needed for task execution. Although motion planning strategies reduce the necessity to specify each minor action, they still necessitate the identification of higher-level actions, such as setting goal locations and sequences of via points. To address these challenges, recent research has pivoted towards Learning from Demonstration (LfD), which is a paradigm in which robots learn new skills by observing and imitating the actions of an expert~\cite{ravichandar2020recent}. However, just as traditional robotics research faces challenges in dataset collection, acquiring demonstration videos for LfD also presents critical difficulties. Despite recent advancements in tools and methodologies designed to simplify data gathering, such as UMI~\cite{chi2024universal}, the process of collecting relevant and comprehensive data continues to be a significant challenge.

The more recent generative model research, such as Large Language Models (LLMs) and Vision Language Models (VLMs), have substantially mitigated this obstacle. These innovations enable robots to operate with pre-trained LLMs and LVMs in a zero-shot manner, such as VoxPoser~\cite{huang2023voxposer}, directly applying human knowledge to finish new robotics tasks without prior explicit training. 

However, prior to the announcement of Sora, applying T2V generative models directly to LfD for robotics was challenging. Most existing models faced difficulties in accurately simulating the physics of complex scenes and often lacked an understanding of specific cause-and-effect instances~\cite{videoworldsimulators2024}. This issue is vital for the effectiveness of robotic learning; discrepancies between the data and real-world conditions can profoundly affect the quality of robot training, compromising the robot's capacity for precise autonomy.

On the other hand, 3D Reconstruction technology, such as Neural Radiance Fields~(NeRF)~\cite{Mildenhall20eccv_nerf} and 3D Gaussian Splatting~(3DGS)~\cite{kerbl20233d}, is getting a lot of attention including in robotics research. DFFs~\cite{shen2023distilled}, capture scene images, extracting dense features via a 2D model and integrating them with a NeRF, mapping spatial and viewing details to create complex 3D models and interactions efficiently. DFFs show that their integration with 3D reconstructed information from NeRF enables robots to better understand and interact with their environment, thus demonstrating how 3D reconstructed information can assist robots in accomplishing tasks more effectively. Apart from this, a notable feature of Sora is its ability to produce multiple shots of the same character within a single sample, ensuring consistent appearance throughout the video~\cite{videoworldsimulators2024}. This capability can be leveraged by integrating 3D reconstruction methods with the multi-shot videos generated by Sora, thereby enhancing robot learning through demonstration by accurately capturing scenes.

\subsection{Infinity 3D Dynamic Scene Reconstruction and Generation}
% 3D science reconstruction achieve significant attentions in most recent. \cite{pollefeys2008detailed} as one of early pointeering work on 3D reconstruction from video to achieve this objective by integrating various technologies such as GPS and inertial measurements, camera pose estimation, and sophisticated algorithms for stereo vision, depth map fusion, and model generation. However, this becomes more simpler after Neural Radiance Fields~(NeRF)~\cite{Mildenhall20eccv_nerf} announced, NeRF allows user direct reconstruct object only from a video. But if want to reconstuct a high-quality science still need more effort on providing as more view as much videos~\cite{li2022neural}. This is even hard to achieved if the object is pretty big or the real enorionment hard to take multi-view videos.

3D scene reconstruction has garnered significant attention in recent years. The work in \cite{pollefeys2008detailed} stands out as one of the pioneering efforts in 3D sense reconstruction from video, achieving remarkable results by integrating various technologies such as GPS, inertial measurements, camera pose estimation, and advanced algorithms for stereo vision, depth map fusion, and model generation. However, the advent of Neural Radiance Fields (NeRF)~\cite{Mildenhall20eccv_nerf} marked a substantial simplification in the field. NeRF enables users to reconstruct objects directly from a video with unprecedented detail. Despite this advancement, reconstructing high-quality scenes still demands significant effort, particularly in acquiring numerous viewpoints or videos, as highlighted in \cite{li2022neural}. This challenge becomes even more pronounced when the object is large, or the environment makes it difficult to capture multi-view videos.

% DyNeRF~\cite{li2022neural} Neural 3D Video Synthesis from Multi-view Video
% 
% Several studies have demonstrated the effectiveness of 3D reconstruction from video, notable examples being DyNeRF~\cite{li2022neural} and NeuralRecon~\cite{sun2021neuralrecon} and existing coherent 3D reconstruction works~\cite{sun2021neuralrecon,wu2024blockfusion} mainly focus on continuously generating new coherent scene blocks by using existing scene's blocks . Meanwhile, Sora~\cite{videoworldsimulators2024}, in particular, showcases its capability to produce multiple views of the same character with physical world simulation ability within a single video. These achievement enables efficient infinity 3D scene reconstruction and generation to become possible, and will widely impact multiple fields. For example, gaming, can directly generate 3D scenes in real time and generate objects that conform to real-world physics without the need for the traditional physics engine required for games.
% 
Several studies have demonstrated the effectiveness of 3D reconstruction from video streams, with notable contributions including DyNeRF~\cite{li2022neural} and NeuralRecon~\cite{sun2021neuralrecon}. Particularly, the forefront of coherent 3D reconstruction research~\cite{sun2021neuralrecon,wu2024blockfusion} is characterized by its emphasis on the seamless generation of new scene blocks, leveraging the spatial and semantic continuity of pre-existing scene structures. Notably, Sora~\cite{videoworldsimulators2024} stands out by offering the capability to render multiple perspectives of a singular character, integrating physical world simulations within a unified video framework. These advancements herald a new era of infinite 3D scene reconstruction and generation, promising substantial implications across various domains. In the realm of gaming, for instance, this technology allows for the real-time generation of 3D environments and the instantiation of objects adhering to real-world physics, potentially obviating the need for conventional game physics engines.

\subsection{Augmented Digital Twins}
% 主要是实时交互和反应

% 在这一框架下，数字孪生的世界中嵌入微分几何这样可以较为精确地描述物理世界的知识，负责绘制出宏观的粗略框架，而大模型根据规则生成实体和个体行为，负责填补更为细致的内容。
% https://www.bilibili.com/read/cv31538259/
Digital twins~(DT) are virtual replicas of physical objects, systems, or processes, designed to simulate the real-world entity in a digital platform~\cite{jones2020characterising}. They are used extensively in various industries for simulation, analysis, and control purposes. The concept is that the DT receives data from its physical counterpart (often in real-time, through sensors and other data-collection methods) and can predict how the physical twin will behave under various conditions or react to changes in its environment. The key aspect of DT is their ability to mirror the physical world accurately, allowing for responses to external signals or changes just as their real-world counterparts would. This capability enables optimization of operations, predictive maintenance, and improved decision-making, as the digital twin can simulate outcomes based on different scenarios without the risk of impacting the physical object.

Sora's world simulation capabilities have the potential to enhance the current DT systems significantly. One of the major challenges in DT is ensuring real-time data accuracy, as unstable network connections can lead to the loss of crucial data segments. Such losses are critical, potentially undermining the system's effectiveness. Current data completion methods often lack an in-depth understanding of the objects' physical properties, relying predominantly on data-driven approaches. By leveraging Sora's proficiency in understanding physical principles, it might be possible to generate data that is physically coherent and more aligned with the underlying real-world phenomena.

Another crucial element of DT is the accuracy with which the visual system responds, effectively simulating and mirroring the real world. Currently, the approach involves users triggering events through programming, followed by the application of machine learning algorithms to predict and subsequently visualize the outcomes. This process is intricate and often lacks a comprehensive understanding of the object's physical characteristics. Implementing Sora could potentially streamline this, enabling unified user interaction with the system. Users could interact directly with the visualization interface created by Sora, which would make real-time predictions and provide accurate and real-world like visual feedback while considering the physical properties of the objects involved.

% \subsection{Large-scale Model Unlearning}
% 隐私性， 视频比文字更隐私敏感

\subsection{Establish Normative Frameworks for AI applications}
With the rapid advancement of large generative models such as DALL·E, Midjourney, and Sora, the capabilities of these models have seen remarkable enhancement.
Although these advancements can improve work efficiency and stimulate personal creativity, concerns about misuse of these technologies also arise, including fake news generation~\cite{chen2023can}, privacy breaches~\cite{liu2023deid}, and ethical dilemmas~\cite{yao2023value}. 
At present, it is becoming necessary and urgent to establish normative frameworks for AI applications.
Such frameworks should answer the following questions legitimately: 

\emph{How to explain the decisions from AI.}
Current AI techniques are mostly regarded as black-box.
However, the decision process of reliable AI should be explainable, which is crucial for supervision adjustment, and trust enhancement.

\emph{How to protect the privacy of users.}
The protection of personal information already is a social concern. With the development of AI, which is a data-hungry domain, more detailed and strict regulations should be established.

\emph{How to ensure fairness and avoid discrimination.}
AI systems should serve all users equitably, and should not exacerbate existing social inequalities. 
Achieving this necessitates the integration of fairness principles, along with the proactive avoidance of bias and discrimination, during the algorithm design and data collection phases.

Overall, normative frameworks will encompass both social and technical dimensions to ensure that AI applications are developed in ways that benefit society as a whole.

\section{Conclusion}
\label{sec:con}
Based on the decomposition from Sora, this survey provides a comprehensive review of current T2V works. 
Concretely, we organized the literature from the perspective of the evolution of generative models, encompassing GAN/VAE, autoregressive-based, and diffusion-based frameworks.
Furthermore, we delved into and reviewed the literature based on three critical qualities that excellent videos should possess: extended duration, superior resolution, and seamless quality. 
Moreover, as Sora is declared to be a real-world simulator, we present a realistic panorama that includes dynamic motion, complex scenes, multiple objects, and rational layouts. Additionally, the commonly utilized datasets and metrics in video generation are categorized according to their source and domain of application. Finally, we identify some remaining challenges and problems in T2V, and also propose potential directions for future development.

\clearpage

% commend in for temp
% \section{Abbreviations List}
% \begin{table*}[!t]
%     \begin{tabular}{@{}llll}
%     \textbf{Symbol} & \textbf{Description} & \textbf{Symbol} & \textbf{Description} \\ % Header
%     % Row 1
%     $G$ & GAN generator & $D$ & GAN discriminator \\
%     $\mathbf{V}(D, G)$ & GAN value function & $x$ & Data \\
%     $p$ & Distribution & $\mathbb{E}(\cdot)$ & Expected value \\
%     $z$ & A uniform or Gaussian distribution & $\mathcal{D}$ & Dataset \\
%     $N$ & Numbers of data samples & $\omega$ & VAE encoder parameters\\
%     $\theta$ & Parameters of model& $\mathcal{L}$ & Loss function\\
%     $v$ & Value & $Q, K, V$ &  Attention query key value\\
%     $\mathcal{W}$ & Weights of model & $b$ & Bias \\
%     $\mathcal{N}$ & Gussion distribution & & \\
%     $\mathcal{V}_{i}$ & $i$-th video & $\mathcal{T}_{i}$& $i$-th description\\
%     $\mathcal{V}^{*}_{j}$ & $j$-th generated video & $\mathcal{T}^{*}_{j}$& the input $j$-th description\\
%     $f_{i}^{k}$ & $k$-th frame of $i$-th video & \\
%     $H$ & height of one frame & $W$ & width of one frame \\
%     & & & \\
%     \end{tabular}
% \end{table*}

% \bibliographystyle{IEEEtran}
\bibliographystyle{ieeetr}
\bibliography{main.bib}
\end{document}